%% file: main.tex
\PassOptionsToPackage{table}{xcolor}
\documentclass[runningheads]{llncs}

\usepackage{eccv}

\usepackage{eccvabbrv}

\usepackage{booktabs} 
\usepackage{xcolor}
\usepackage[normalem]{ulem}

\newcommand{\sysname}{StoryTeller}
\newcommand{\benchname}{StoryAD-QA}
\usepackage{enumitem}

\usepackage{booktabs}
\usepackage{subcaption}
\usepackage{siunitx}

\usepackage{graphicx}
\usepackage{booktabs}
\usepackage{tcolorbox}
\tcbuselibrary{breakable}
\tcbset{
  colback=gray!5,
  colframe=gray!60,
  boxrule=0.4pt,
  arc=2pt,
  left=4pt,
  right=4pt,
  top=4pt,
  bottom=4pt,
  breakable
}
\usepackage{iftex}
\ifPDFTeX
  \usepackage[accsupp]{axessibility}  
\fi

\usepackage{hyperref}

\usepackage{pifont}

\begin{document}

\title{StoryTeller: Training-Free Narrative Grounding for Long-Form Audio Description}
\titlerunning{StoryTeller}

\author{Seung Hyun Hahm \and Minh T. Dinh \and SouYoung Jin}
\authorrunning{S.~H.~Hahm et al.\ }

\institute{Dartmouth College, USA\\
\email{\{Seung.Hyun.Hahm.GR,Minh.T.Dinh.GR,SouYoung.Jin\}@dartmouth.edu}}

\maketitle

\begin{abstract}
Long-form audio description (AD) requires more than describing visible actions: it must preserve characters, events, relationships, and story context across scenes so that blind and low-vision (BLV) audiences can follow a film. Modern video--language models (VLMs) are effective on short clips, but they often treat each moment independently, producing descriptions that miss who characters are, why events matter, and how the current scene connects to earlier narrative context. We propose \sysname{}, a training-free framework for story-aware long-form AD. Instead of relying only on local visual cues, \sysname{} maintains a verified narrative memory that carries forward story-relevant information across scenes, enabling later descriptions to remain coherent, grounded, and contextually informative. Given only raw video and a movie title, \sysname{} can optionally retrieve public movie metadata to resolve names and story context, while accepting only facts that are supported by the video through semantic filtering and VLM verification. The method requires no subtitles, scripts, AD transcripts, aligned captions, character banks, precomputed face identities, or task-specific fine-tuning. To evaluate whether generated AD preserves narrative information, we introduce \benchname{}\footnote{\textbf{\benchname{} benchmark dataset and evaluation code:}\\\url{https://github.com/SEE-AI-Lab/ECCV2026_StoryTeller_StoryAD_QA}}, a question-answering benchmark that tests whether a language model can answer story-context questions using only the generated descriptions. Experiments on standard AD benchmarks and diverse long-form videos show that \sysname{} consistently improves narrative coherence, factual grounding, and story comprehension over strong baselines in automatic, QA-based, and human evaluations.
\end{abstract}

\input{sec/1_intro}
\input{sec/2_related_work}
\input{sec/3_proposed_approach}
\input{sec/4_benchmark}
\input{sec/5_experiments}
\input{sec/6_conclusion}

\bibliographystyle{splncs04}
\bibliography{main}
\end{document}


\title{StoryTeller: Training-Free Narrative Grounding for Long-Form Audio Description \\
--Supplementary Material--}

\titlerunning{StoryTeller}

\author{Seung Hyun Hahm \and Minh T. Dinh \and SouYoung Jin}
\authorrunning{S.~H.~Hahm et al.}
\institute{Dartmouth College}



\maketitle

\input{Supplementary_Material/supp_src_codes/sec/7_supple}

\bibliographystyle{splncs04}
\bibliography{main}

%% file: sec/1_intro.tex
\section{Introduction}
\label{sec:1_intro}

\begin{figure}[t]
\centering
\includegraphics[width=\linewidth]{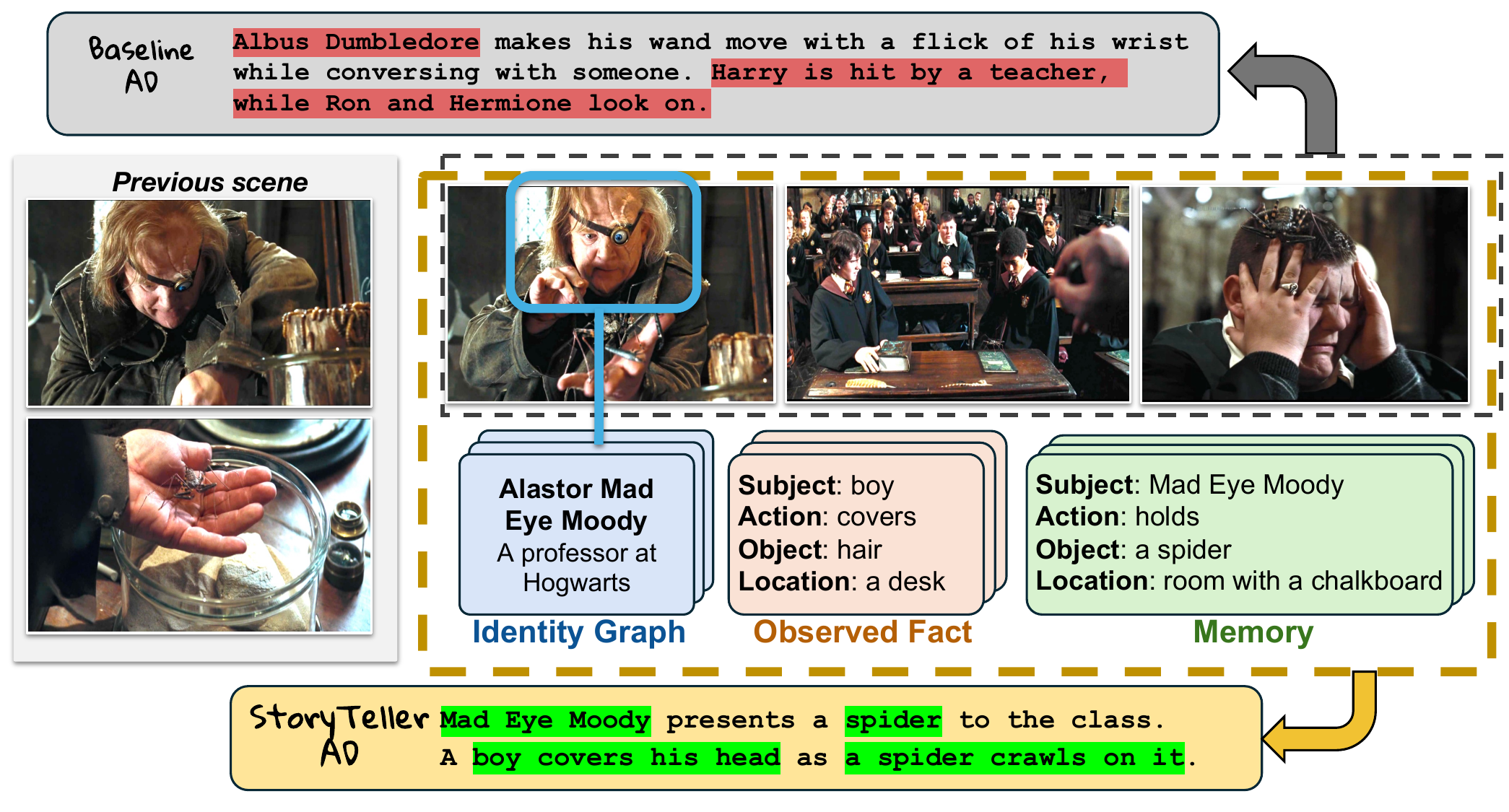}
\vspace{-0.2in}
\caption{\textbf{Story-level audio description requires narrative memory.}
Existing audio description (AD) systems typically generate descriptions using only local visual context or static character banks, which often fails to preserve long-range narrative relationships. 
Our \sysname{} instead maintains a persistent narrative state consisting of an identity graph that tracks characters across scenes and a salience-weighted memory that accumulates narrative facts over time. 
This enables consistent character references and long-range story reasoning across an entire film. The baseline example is the output of AutoAD-Zero.
}
\vspace{-0.2in}
\label{fig:teaser}
\end{figure}

Audio description (AD) enables blind and low-vision (BLV) audiences to access visual media by narrating important visual information between dialogue and other sounds. Effective AD goes beyond describing objects or actions in the current frame: it identifies who is present, explains what has just happened, conveys why a moment matters, and connects the scene to the broader narrative. Professional AD writers are able to provide this narrative grounding because they watch the film and write with the whole story in mind. However, producing high-quality AD remains costly and time-intensive, and copyright restrictions make large publicly distributable AD datasets difficult to obtain.

Recent advances in video--language models (VLMs)\cite{singh2025openai, li2026qwen3, wang2025internvl3, videollama2} and large language models have motivated research on automatic AD generation. State-of-the-art identity-aware AD systems
~\cite{autoad1,autoad2,raajesh2024micap}
are typically trained as task-specific models on curated movie datasets such as LSMDC~\cite{rohrbach2017movie} and MAD~\cite{Soldan_2022_CVPR}. These pipelines incorporate identity-aware components built from face-recognition backbones trained on labeled data. While effective in benchmark settings, this reliance on copyrighted training data, curated movie annotations, and specialized identity modules limits their scalability, reproducibility, and applicability to new films.

Training-free pipelines, including AutoAD-Zero~\cite{autoadzero} and MM\_Narrator~\cite{zhang2024mm}, have recently emerged as an alternative to supervised AD generation. These methods remove the need for task-specific model training, but they often still rely on resources prepared before AD generation, including character banks, aligned subtitles, or existing AD transcripts. Moreover, they typically propagate context in shallow or static forms, such as concatenating previous descriptions or reusing fixed identity features. Although these strategies can improve local consistency across nearby clips, they do not explicitly model how narrative information evolves over time: which characters, relationships, and events should remain salient, and which details should fade as the film progresses.

Human viewers do not understand a film by retaining a complete list of previous frames. Instead, they maintain an evolving representation of the story, where central characters, unresolved conflicts, and recurring relationships remain salient, while incidental visual details gradually lose importance. This perspective suggests that long-form AD should preserve identity continuity, causal structure, and evolving character interactions across a film, without relying on external resources such as subtitles, scripts, AD transcripts, character banks, or precomputed face identities.

We introduce \textbf{\sysname{}}, a training-free framework for long-form AD that maintains an explicit story state as it processes clips in chronological order. Given a sequence of video clips $\{v_1, \dots, v_T\}$, the system keeps two forms of memory: an identity graph for recurring characters and a salience-weighted memory for verified story facts. Each new clip updates this state through three steps: track visible identities, extract candidate facts with optional public movie metadata, and verify those facts against the video before updating memory. The memory then reinforces facts that remain relevant and decays facts that no longer matter.

\sysname{} takes as input only the raw video and a movie title. The title may be used to retrieve public IMDb plot summaries, character lists, and short character descriptions. This retrieved text is only a source of possible names or context; it is not copied into the AD, stored directly as memory, or treated as ground truth. A fact enters memory only if it is supported by the current clip and accepted by semantic filtering and VLM verification. 

Beyond generation, we also address evaluation. Standard metrics such as BLEU~\cite{papineni2002bleu}, CIDEr~\cite{vedantam2015cider}, and SPICE~\cite{anderson2016spice} measure lexical overlap between generated descriptions and reference captions, but they do not evaluate whether a listener can reconstruct the narrative from the AD alone. To address this limitation, we introduce~\textbf{\benchname{}}, a narrative-comprehension question-answering benchmark. For each scene, we generate questions whose answers depend on both the current clip and preceding narrative context. During evaluation, a language model receives only the generated audio descriptions and must answer these questions without access to the video. This protocol directly measures whether the description preserves characters, events, and causal relationships necessary for story understanding rather than merely describing visual details in isolation.

\subsection*{Contributions.}
\begin{itemize}
    \item We propose \sysname{}, a fully training-free framework for long-form audio description that uses raw video and public title-keyed metadata without subtitles, AD transcripts, manually curated visual character banks, or task-specific fine-tuning.
    \item We introduce an explicit narrative memory mechanism with reinforcement--decay dynamics that approximates how human viewers track evolving story salience across scenes.
    \item We present \benchname{}, a narrative-comprehension QA benchmark that evaluates whether generated audio descriptions support story understanding beyond lexical similarity.
\end{itemize}

%% file: sec/2_related_work.tex
\section{Related Work}
\label{sec:related_work}

\noindent\textbf{Benchmarking Audio Description Generation.}
Movie understanding has often been studied as question answering over long videos~\cite{tapaswi2016movieqa, vicol2018moviegraphs, huang2020movienet, song2024moviechat, yue2023movie101}. For example, MovieQA~\cite{tapaswi2016movieqa} uses questions grounded in plot summaries, scripts, and subtitles, while~\cite{vicol2018moviegraphs} and~\cite{huang2020movienet} annotate characters, interactions, and story structure. MovieChat~\cite{song2024moviechat} emphasizes on long-video QA with sparse memory and introduces MovieChat-1K. These datasets are crucial for evaluating video reasoning, but they do not directly ask whether an AD gives a listener enough information to understand the story without seeing the video.

Datasets designed specifically for AD instead focus on localized narration. LSMDC~\cite{rohrbach2017movie} and MAD~\cite{Soldan_2022_CVPR} provide large-scale movie captioning resources aligned to short clips, and recent AD generation systems~\cite{autoad1,autoad2,autoad3,raajesh2024micap,autoadzero,park2025narrad,zhang2024mm}
build on these datasets with identity-aware modules or curated character banks. Despite their usefulness, evaluation in these works often relies on direct string comparison with professional AD using metrics such as CIDEr~\cite{vedantam2015cider}, SPICE~\cite{anderson2016spice}, and ROUGE~\cite{linrouge}, which are highly sensitive to temporal alignment. Addressing this limitation, AutoAD~II~\cite{autoad2} introduces an evaluation protocol that measures whether generated descriptions are semantically aligned with nearby ground-truth annotations rather than requiring exact sentence matches. Building on this direction, AutoAD~III~\cite{autoad3} further proposes CRITIC for evaluating character retrieval and LLM-AD-Eval for sentence-level semantic scoring with LLMs.

More recently, question-answering evaluation has been explored to assess the usefulness of AD. ADQA~\cite{kala2025adqa} evaluates whether generated AD supports visual appreciation and narrative understanding in a certain video segment via questions generated from reference descriptions. Instead, our benchmark focuses on long-form narrative grounding, evaluating whether ADs preserve information required for reasoning over story context that spans multiple scenes.

\noindent\textbf{Identity Modeling in Video Narratives.}
Maintaining consistent character identity across scenes is essential for long-form video understanding and AD. Several recent AD systems explicitly rely on \emph{character banks}--precomputed repositories of character-specific visual features, e.g., face embeddings or reference images used to recognize recurring characters throughout a movie~\cite{autoad1,autoad2,autoad3}. In contrast, \sysname{} requires no character bank or precomputed face identities. Instead, it discovers recurring identities directly from repeated face observations, allowing character identities to emerge dynamically as narrative evidence accumulates.

\noindent\textbf{Retrieval-Augmented Generation.}
Retrieval-augmented generation (RAG) improves grounding by incorporating external evidence during generation~\cite{lewis2021retrievalaugmentedgenerationknowledgeintensivenlp,gao2023rag_survey}. However, naive retrieval can introduce irrelevant or contradictory information and may amplify hallucinations in multimodal models~\cite{huang2024hallucination,shao2024visual}. Prior work improves retrieval quality through dense retrievers and structured fusion mechanisms~\cite{guu2020realm,karpukhin2020dpr,izacard2021fid}, but these approaches focus primarily on factual knowledge grounding rather than narrative coherence. Our approach instead performs narrative-aware retrieval restricted to movie-scoped sources and triggered only when scene observations indicate potentially story-relevant events.

\noindent\textbf{Memory for Long-Horizon Reasoning.}
Memory mechanisms have been widely explored for long-horizon multimodal reasoning. Recent systems such as VideoAgent~\cite{fan2024videoagent}, Optimus-1~\cite{li2024optimus}, and M3-Agent~\cite{long2025m3agent} maintain explicit temporal memories to support extended video reasoning. Other long-video systems employ episodic memories for VideoQA~\cite{wang2025videoem}, retrieval-augmented memories for long-video comprehension~\cite{luo2024videorag}, or global audio-visual character representations for dense video description~\cite{he2024storyteller}. These methods primarily target long-video QA, embodied reasoning, video comprehension, or dense video description. In contrast, \sysname{} is designed for accessibility-oriented audio description, where generated descriptions are intended to complement the original movie audio rather than summarize the entire film. Instead of maintaining a general-purpose memory, \sysname{} stores a lightweight narrative memory of visually verified story facts with evolving salience weights, allowing important characters and events to persist across scenes while transient details gradually fade.

%% file: sec/3_proposed_approach.tex
\section{\sysname}
\label{sec/3_proposed_approach}

\begin{figure}[t!]
    \centering
    \includegraphics[width=0.9\linewidth, trim=0pt 40pt 0pt 100pt, clip,]{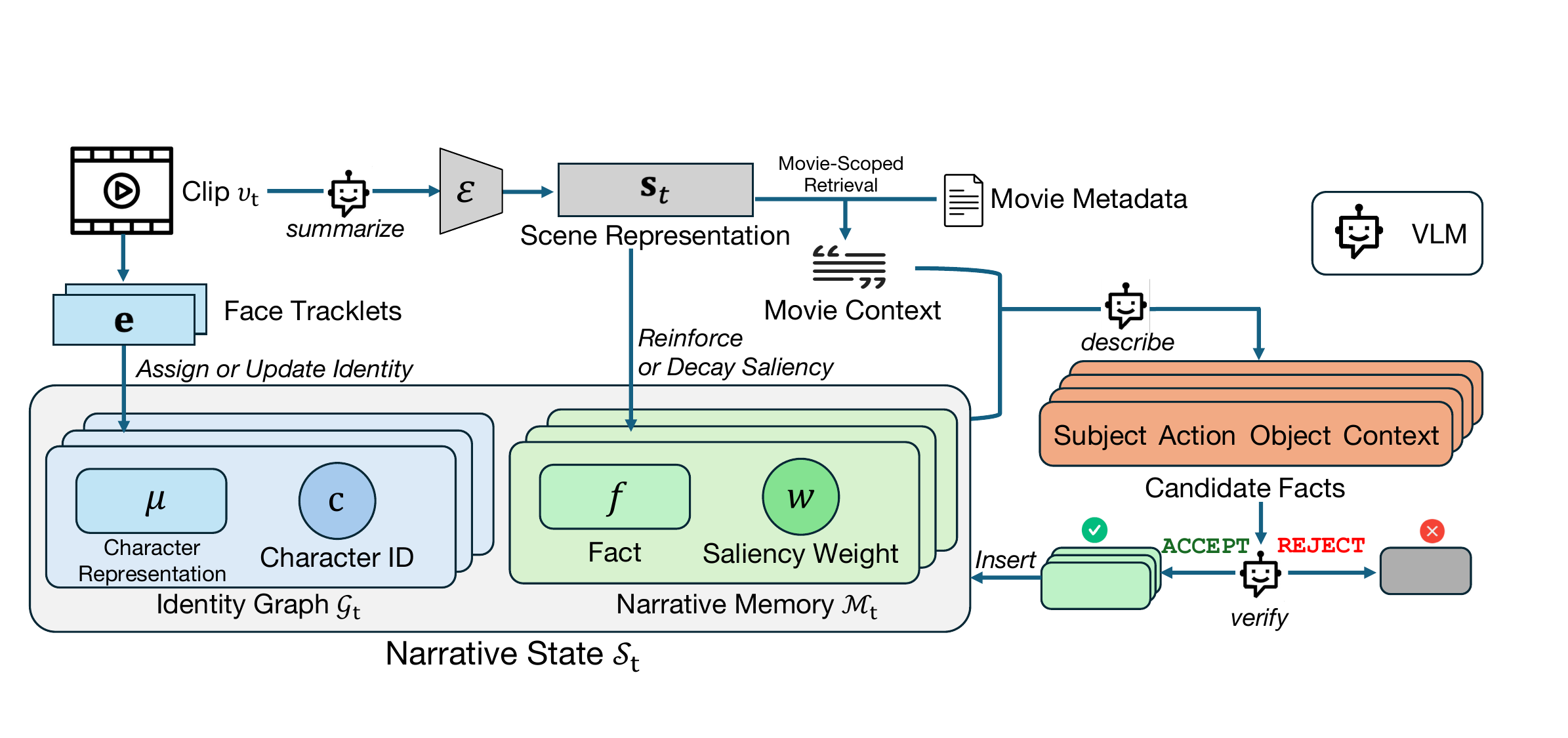}\\
    \vspace{-0.1in}
    \caption{\textbf{Overview of \sysname{}.}
Before processing clip $v_t$, the system maintains the narrative state $S_{t-1}=(\mathcal{G}_{t-1},\mathcal{M}_{t-1})$, where $\mathcal{G}_{t-1}$ is the identity graph and $\mathcal{M}_{t-1}$ is the narrative memory. Given $v_t$, StoryTeller (1) updates the identity graph to maintain character continuity, (2) extracts structured factual candidates using the updated identity graph, optional public movie metadata, and the previous narrative memory, and (3) updates the narrative memory using semantic filtering, VLM verification, and reinforcement--decay salience dynamics. Public movie metadata may suggest character names or plot context, but only visually verified facts are stored in narrative memory.
}
    \vspace{-0.2in}
    \label{fig:overview}
\end{figure}

Figure~\ref{fig:overview} illustrates the overall pipeline of \sysname{}. The input is a movie divided into chronological \textit{clips} $\{v_1, \dots, v_T\}$. \sysname{} processes the clips sequentially while carrying a persistent \emph{narrative state} that summarizes story information accumulated from previously processed clips. This state enables the system to track recurring characters, extract scene-level events, and preserve long-range narrative context throughout the movie.

Before processing {clip $v_{t}$}, the system maintains the narrative state

\begin{equation}
\mathcal{S}_{t-1} = (\mathcal{G}_{t-1}, \mathcal{M}_{t-1}),
\end{equation}

where $\mathcal{G}_{t-1}$ is the identity graph storing visual representations of recurring characters, and $\mathcal{M}_{t-1}$ is the narrative memory storing verified story facts together with their salience weights.

Processing clip $v_t$ updates the narrative state according to

\begin{equation}
\mathcal{S}_{t} = \Phi(v_t, \mathcal{S}_{t-1}),
\end{equation}

where $\Phi$ denotes the overall state update operator.

For each new clip, \sysname{} updates its narrative state through three modules. First, the \emph{identity graph update} links visible faces to existing character nodes when there is sufficient visual evidence, allowing the system to maintain consistent references to recurring characters across clips. Second, \emph{grounded fact induction} proposes structured facts about the current scene, optionally using public movie metadata as hints for names or story context. These candidate facts are then filtered to ensure that they are semantically relevant to the scene and supported by the video. Third, the \emph{salience-based narrative memory} updates the importance of stored facts: facts that remain relevant are reinforced, while facts that no longer matter gradually decay. This allows important characters, events, and relationships to persist across scenes without carrying forward every transient detail.

Besides the visual clips, the movie title is used to retrieve optional public movie metadata. The retrieved metadata may suggest character names or plot context, but it cannot by itself create a memory entry or directly influence the generated narration. A fact is added to the narrative memory only if it is supported by the current clip and accepted by the semantic filtering and VLM verification steps.
\subsection{Identity Graph Update}
\label{sec:identity}

The identity graph {$\mathcal{G}_{t-1}$} stores character representations accumulated from previously processed clips. Its role is to associate recurring appearances of the same person across clips, fostering consistent character references throughout the movie.

Unlike conventional character banks in the AD literature~\cite{autoad3,autoadzero}, which store static character representations, our identity graph is updated continuously as new visual observations and verified narrative evidence become available. This allows recurring characters to be tracked consistently across scenes, while novel character identities can be established progressively as sufficient supporting evidence accumulates. 

\noindent\textbf{Tracklets and Embeddings.}
For each clip $v_t$, faces are detected in each frame and associated into short temporal tracklets $\mathcal{T}_k$, where each tracklet represents a single face observed over consecutive frames. A pretrained ArcFace encoder $\phi$~\cite{deng2019arcface} extracts an embedding $\mathbf{v}_{t,i}=\phi(x_{t,i})$ for each face crop $x_{t,i}$. The embedding of tracklet $\mathcal{T}_k$ is obtained by averaging the embeddings of all face observations in the tracklet:
\begin{equation}
\mathbf{e}_k
=
\frac{1}{|\mathcal{T}_k|}
\sum_{i\in\mathcal{T}_k}
\mathbf{v}_{t,i}.
\end{equation}

\noindent\textbf{Identity Graph Representation.}
The identity graph at time $t$ is defined as

\begin{equation}
\mathcal{G}_t =
\{(c_j, \boldsymbol{\mu}_j)\}_{j=1}^{N_t},
\end{equation}
where $c_j$ denotes a character identity and $\boldsymbol{\mu}_j$ is the mean embedding of all tracklets assigned to that character. The graph therefore maintains a compact representation of characters observed up to time $t$.

\noindent\textbf{Identity Assignment.}
For each new tracklet embedding $\mathbf{e}_k$, we compute its cosine similarity to every existing identity embedding:

\begin{equation}
j^* =
\arg\max_j
\cos(\mathbf{e}_k,\boldsymbol{\mu}_j).
\end{equation}

If $\cos(\mathbf{e}_k,\boldsymbol{\mu}_{j^*}) > \tau$, the tracklet is assigned to identity $c_{j^*}$. Otherwise, a new identity node is created and the tracklet remains unnamed until sufficient evidence is accumulated to associate it with a character.

\noindent\textbf{Identity Reseeding.}
Characters often appear before their names are revealed. Accordingly, every face tracklet is initially represented as an anonymous identity. The VLM may consult optional public movie metadata when proposing a character name, but the current clip must provide sufficient visual and narrative evidence before the association is accepted. Once verified, the corresponding anonymous identity is assigned the character name, and the identity graph is updated by creating a new named node or updating an existing one. Previously observed or future tracklets that are visually similar can then be associated with the same identity. Thus, the similarity threshold $\tau$ serves only as a candidate matching criterion rather than proof of identity. Tracklets that lack sufficient visual or narrative evidence remain anonymous until additional evidence becomes available.

\subsection{Grounded Fact Induction}
\label{sec:fact_induction}

Given the current clip $v_t$, the {updated} identity graph $\mathcal{G}_{t}$ containing recurring character identities, and the narrative memory $\mathcal{M}_{t-1}$ summarizing previously accumulated story information, \sysname{} then inducts facts, i.e., extracts structured factual candidates and filters them before they are inserted into memory. The objective is to identify facts that are both visually supported and consistent with the evolving narrative context to achieve a new  narrative memory $\mathcal{M}_{t}$.

\noindent\textbf{Scene Representation.}
We first summarize the visual content of the clip. A VLM generates a concise scene summary $s_t$ describing the primary event in $v_t$. The summary is then embedded as $\mathbf{s}_t=\varepsilon(s_t)$, where $\varepsilon(\cdot)$ denotes the text embedding function.

\noindent\textbf{Public Movie Metadata Retrieval.}
Before fact induction, \sysname{} uses the movie title to retrieve optional public movie metadata, including plot summaries, character lists, and short character descriptions from IMDb. Each paragraph is embedded, and the paragraphs most similar to the current scene summary $\mathbf{s}_t$ are retrieved. The retrieved metadata serves only as auxiliary context: it may suggest character names or plot context, but it is never copied into the narration or directly stored in the narrative memory. All stored facts must be supported by the current clip and accepted by the semantic filtering and VLM verification steps.

If little or no metadata is available, retrieval simply returns fewer or no passages. The remainder of the pipeline is unchanged: facts are still extracted and verified, the identity graph and narrative memory are updated from visual evidence, and unnamed characters remain anonymous until sufficient evidence supports an identity assignment. 

\noindent\textbf{Structured Fact Extraction.}
Given clip $v_t$, the VLM proposes structured factual candidates conditioned on three sources of context: (i) recurring character identities from the identity graph $\mathcal{G}_{t}$, (ii) optional public movie metadata, and (iii) relevant narrative context retrieved from the memory $\mathcal{M}_{t-1}$. Each candidate is represented as
\[
f=(\text{subject},\text{action},\text{object},\text{context}),
\]
where the subject may correspond to either a named character or an anonymous identity. The extracted candidates describe potential narrative updates, including character actions, interactions, object states, and scene context.

This stage is intentionally permissive: it generates hypotheses rather than committing facts to memory. Candidate facts, including proposed character names inferred from optional public movie metadata, are accepted only after the semantic filtering and VLM verification.

\noindent\textbf{Semantic Filtering.}
The extracted candidates may include facts that are only weakly related to the current clip or inconsistent with the evolving narrative. Before applying computationally expensive VLM verification, we use a lightweight semantic filter to retain candidates that are either semantically related to the current scene or consistent with previously verified facts.

Each candidate fact $f$ is embedded as $\mathbf{z}_f=\varepsilon(f)$. We compute
\begin{equation}
g_f =
\max
\left(
\cos(\mathbf{z}_f,\mathbf{s}_t),
\max_m \cos(\mathbf{z}_f,\varepsilon(f_m))
\right),
\end{equation}
where $f_m$ denotes the $m$-th fact stored in the narrative memory $\mathcal{M}_{t-1}$. The first term measures semantic similarity to the current scene summary, while the second measures consistency with previously verified facts. Candidate facts with $g_f>\sigma$ are forwarded to VLM verification. We use a conservative high-recall threshold for $\sigma$ to discard only clearly unrelated candidates while retaining plausible facts for subsequent verification.

\noindent\textbf{Verification Stage.}
Candidates that pass semantic filtering are evaluated by a dedicated VLM verifier using the current clip. The verifier outputs \texttt{ACCEPT} or \texttt{REJECT}. Only accepted facts are inserted into the narrative memory or used to update named identities. This stage prevents semantic consistency from being mistaken for visual evidence: a candidate may agree with retrieved metadata or previously verified memory, but it cannot update the identity graph or narrative memory unless the video supports it.
\subsection{Salience-Based Narrative Memory}
\label{sec:memory}

After verification, accepted facts are inserted into the narrative memory $\mathcal{M}_t$. Each memory entry consists of a validated fact and an associated salience weight that determines its influence on future narration.

Formally, the memory at time $t$ is
\begin{equation}
\mathcal{M}_t =
\{ (f_m, w_m^{(t)}) \},
\end{equation} where $f_m$ is the $m$-th validated fact and $w_m^{(t)}$ is its salience weight at time $t$. Accepted facts are inserted into the narrative memory with an initial salience weight $w_m^{(0)}=0.25$.

\noindent\textbf{Relevance Computation.}
For the current scene summary $s_t$, we compute the semantic similarity between the scene embedding $\mathbf{s}_t$ and each stored fact embedding $\mathbf{z}_m=\varepsilon(f_m)$:

\begin{equation}
r_m^{(t)} = 
\cos(\mathbf{s}_t,\mathbf{z}_m).
\end{equation}

\noindent\textbf{Reinforcement--Decay Dynamics.}
The relevance score $r_m^{(t)}$ determines how strongly fact $f_m$ should be reinforced. Memory weights evolve through reinforcement and decay: at each scene, every fact decays slightly, and facts relevant to the current scene are reinforced in proportion to their relevance:
\begin{equation}
w_m^{(t)} = \lambda\, w_m^{(t-1)} + \alpha\, r_m^{(t)},
\qquad \lambda \in (0,1).
\end{equation}
Here $\lambda$ is the decay factor, applied to the previous weight, and $\alpha$ is the reinforcement strength. A fact that stays relevant across scenes keeps gaining weight and remains salient, while a fact that stops being relevant ($r_m^{(t)}=0$) loses a fixed fraction of its weight each scene and gradually fades. Supplementary Sec.~S5 visualizes persistent and transient memory dynamics.

%% file: sec/4_benchmark.tex
\section{\benchname: Evaluating Narrative Comprehension} 
\label{sec:benchmark}
Existing captioning metrics such as CIDEr~\cite{vedantam2015cider}, SPICE~\cite{anderson2016spice}, ROUGE~\cite{linrouge}, and BLEU~\cite{papineni2002bleu} measure lexical or semantic similarity between generated and reference descriptions. While effective for evaluating caption quality at the scene level, they do not assess whether a sequence of audio descriptions preserves the information needed to understand a story. Narrative comprehension requires tracking characters, relationships, events, and causal developments across scenes, yet ADs are essentially concise and may omit important context. As a result, high captioning scores do not necessarily indicate strong story-level understanding. To address this limitation, we introduce \benchname{}, a multiple-choice benchmark that evaluates whether generated audio descriptions retain sufficient information for downstream narrative reasoning. Rather than comparing generated descriptions against reference text, \benchname{} measures whether questions about the story can be answered using only the generated AD.
\begin{figure*}[bh!]
\vspace{-0.15in}
\centering
\includegraphics[width=0.9\linewidth, trim=0pt 40pt 0pt 10pt, clip]{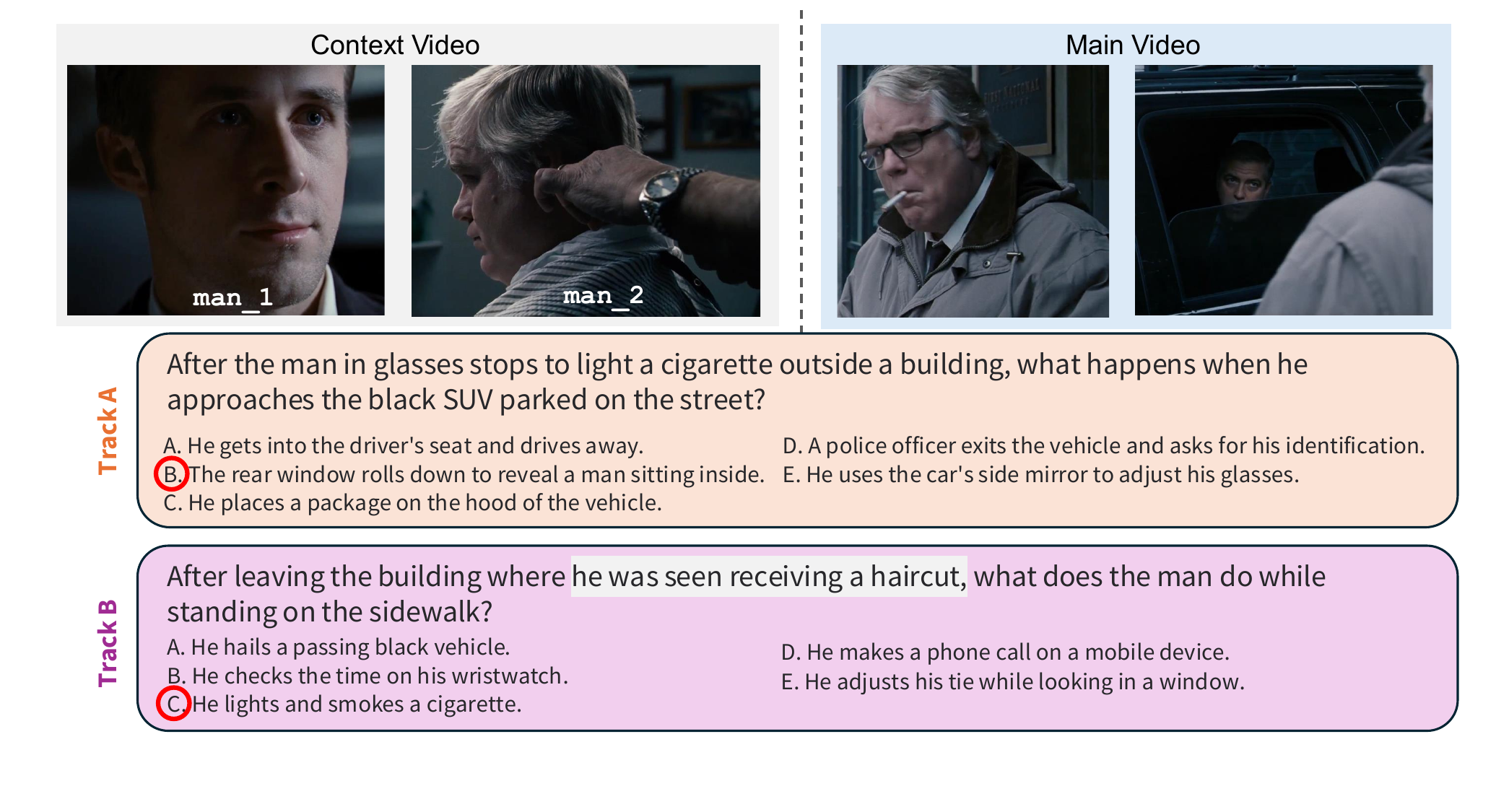}
\vspace{-0.1in}
\caption{\textbf{Example of questions from two tracks in \benchname.}
Grey panels show frames from the preceding context window used in Track~B for question generation. The context reveals that the man was receiving a haircut inside the building, enabling the question to reference this earlier event. Blue panels show the main clip, where the man stands outside the building and lights a cigarette, which determines the correct answer. During evaluation, only the AD of the main clip is provided, so the description must preserve or reintroduce the earlier context.
}
\label{fig:bench_examples}
\end{figure*}
\subsection{Benchmark Construction}
\benchname\ is constructed from movie clips derived from the MAD-Eval~\cite{Soldan_2022_CVPR} dataset. 
In total, the benchmark contains \textbf{2,574 questions} across two evaluation tracks: 1,611 in Track A (segment-only QA) and 963 in Track B (context-conditioned QA). See Supplementary Table~S6 for additional dataset statistics.

Each example in \benchname\ consists of a video segment and, following prior work~\cite{kala2025adqa,lei2018tvqa}, a multiple-choice question with five answer options, exactly one of which is correct. Questions are generated automatically using a vision-language model (VLM) that observes the original video along with short movie-level metadata describing the overall story context. The VLM is instructed to focus on clearly observable events or outcomes and to avoid relying on dialogue, external knowledge, or subjective interpretation. This ensures that answers are grounded in visual evidence rather than speculation. All answer choices are randomly shuffled before evaluation to reduce positional bias. {During evaluation, the answering model receives only the AD text for the evaluated segment and the answer choices; it receives no video, retrieved IMDb/Wikipedia metadata, movie title, or dialogue transcript.

\subsection{Evaluation Protocol}
To evaluate whether generated AD preserves narrative information beyond individual frames, \benchname{} uses multiple-choice question answering over continuous movie segments. Questions are generated from video segments, but during evaluation an \emph{answerer} (a Language Model) receives only the generated AD text, together with the question and answer choices, without access to the original video. Accuracy is the fraction of questions answered correctly from the AD text alone. We report two complementary evaluation tracks (See Table~\ref{tab:storyqa}).

\noindent\textbf{Track A: Segment-only QA.}
In Track A, questions are generated from a single continuous video segment of 30, 60, 120, or 240 seconds. During evaluation, the answering model receives the generated ADs for that same segment and selects the correct option. This track asks whether the AD captures the key events, entities, and relationships within the observed segment.

\noindent\textbf{Track B: Context-conditioned QA.}
In Track B, each question is constructed from a fixed 30-second main clip together with its preceding context. The question generator observes the main clip plus $c \in {30,60,90}$ seconds of video immediately before it. During evaluation, however, the answerer receives only the AD generated for the 30-second main clip, along with the question and answer choices; it does not receive the preceding context or the original video. This setting tests whether the main-clip AD preserves or reintroduces contextual information from earlier scenes, such as character identities, ongoing actions, and narrative relationships.

All questions retained in \benchname{} are manually verified before inclusion. Human annotators check that each question is visually grounded, has plausible distractors, and has a correct answer. Low-quality questions are regenerated and re-verified. Additional details are provided in Supplementary Sections~S8.2--S8.3.

%% file: sec/5_experiments.tex
\section{Experiments}
\label{sec:experiments}

We evaluate \sysname{} along three dimensions:
(i) captioning quality on MAD-Eval, (ii) narrative comprehension with the proposed \benchname{} benchmark, and (iii) the contribution of individual pipeline components through ablation studies. All experiments are conducted without task-specific training, subtitles, scripts, AD transcripts, character banks, or precomputed face identities.

\subsection{Experimental Setup}

\noindent\textbf{Datasets.}
We evaluate our method on MAD-Eval~\cite{Soldan_2022_CVPR}, which consists of 10 movies from the LSMDC~\cite{rohrbach2017movie} dataset. LSMDC contains 202 movies paired with professionally authored audio descriptions aligned to the corresponding video clips. To evaluate long-range narrative understanding, we further introduce and evaluate on \benchname{}, a question-answering benchmark constructed from non-overlapping segments of the MAD-Eval movies.

\noindent\textbf{Implementation Details.}
We use Qwen3-VL-2B-Instruct~\cite{li2026qwen3} for scene summarization and structured fact extraction. We use Qwen3-VL-8B-Thinking for fact verification, which outputs an ACCEPT/REJECT decision for each candidate fact. Text embeddings are computed with Qwen3-VL-Embedding-2B~\cite{li2026qwen3}, and face embeddings are extracted with ArcFace~\cite{deng2019arcface}. Gemini3-Flash~\cite{Gemini3Flash} is used only for \benchname{} question generation and answering. For question generation, Gemini3-Flash receives the video and a short Wikipedia overview. For answering, it receives only the AD text, question, and answer choices. IMDb retrieval is used only by the AD-generation pipeline and is defined in Section~\ref{sec:fact_induction}. Additional implementation details are provided in Supplementary Sec.~S4, and the full prompts are provided in Supplementary Sec.~S10.

\noindent\textbf{Threshold Calibration.}
Identity assignment uses a fixed cosine similarity threshold of $\tau=0.58$ for ArcFace embeddings. The semantic grounding threshold is set to $\sigma=0.20$, calibrated from grounding-score statistics collected on separate calibration runs to provide a high-recall semantic filter that removes only clearly unrelated candidates before VLM verification. Both thresholds are fixed across all experiments and are not tuned on the evaluation data. Additional calibration statistics and threshold analyses are provided in Supplementary Sec.~S4.2.

\noindent\textbf{Runtime.}
On one A100-SXM4-80GB GPU, StoryTeller processes a feature-length movie in approximately 3--4 hours. The semantic filter is approximately $3{,}000\times$ faster than VLM verification, substantially reducing unnecessary verifier calls. 

\noindent\textbf{Memory Parameters.}
Narrative memory uses the reinforcement--decay mechanism described in Sec.\ref{sec:memory}. All hyperparameters are fixed across experiments; complete values are provided in Supplementary Table~S4.3.

\subsection{MAD-Eval Benchmark Comparison}

Table~\ref{tab:mad_eval} compares StoryTeller with prior audio description systems while separating required resources from captioning performance. We additionally evaluate a variant without public movie metadata retrieval in the ablation study (Table~\ref{tab:ablation}).

\begin{table}[t]
\centering
\caption{
\textbf{Comparison on MAD-Eval.}
\textit{Train} denotes whether task-specific training is required.
\textit{Curated Res.} includes subtitles, scripts, AD transcripts, aligned captions, character banks, or precomputed face identities. \textit{Public Meta.} means public movie text retrieved by title, such as IMDb plot summaries, character lists, and short character descriptions.
\textit{VLM} indicates the underlying vision-language model used by each method.
Our method requires neither additional training nor curated or precomputed movie resources; the default setting uses public movie metadata.
}
\vspace{-0.1in}
\label{tab:mad_eval}
\setlength{\tabcolsep}{5pt}

\resizebox{\linewidth}{!}{%
\begin{tabular}{lcccc ccc}
\toprule
& \multicolumn{4}{c}{Method Setup} & \multicolumn{3}{c}{Metrics} \\
\cmidrule(lr){2-5} \cmidrule(lr){6-8}
Model & Train & Curated Res. & Public Meta. & VLM & CIDEr & SPICE & ROUGE-L \\
\midrule
AutoAD-I~\cite{autoad1}   & \checkmark & \checkmark & $\times$ & - & 14.3 & 4.4 & 11.9 \\
AutoAD-II~\cite{autoad2}  & \checkmark & \checkmark & $\times$ & - & 19.2 & - & 13.4 \\
AutoAD-III~\cite{autoad3} & \checkmark & \checkmark & $\times$ & - & \textbf{24.0}& - & - \\
\midrule
MM-Vid~\cite{2023mmvid}      & $\times$ & \checkmark & \checkmark & GPT-4V & 6.1 & 6.1 & 9.8 \\
MM-Narrator~\cite{zhang2024mm} & $\times$ & \checkmark & \checkmark & GPT 4 & 13.9 & 5.2 & 13.4 \\
AutoAD-Zero~\cite{autoadzero} & $\times$ & \checkmark & \checkmark & VideoLLaMA2 & 22.4& 7.3 & 14.4 \\
\midrule
\textbf{\sysname{}} & $\times$ & $\times$ & \checkmark & VideoLLaMA2 & 19.1 & \textbf{9.0} & \textbf{16.0} \\
\textbf{\sysname{}} & $\times$ & $\times$ & \checkmark & Qwen3-VL & 21.4& 6.7& 15.3\\
\bottomrule
\end{tabular}
}

\end{table}

We additionally evaluate StoryTeller on MovieChat~\cite{song2024moviechat} using 9 videos and 27 questions. Results are reported in Supplementary Sec.~S7.

\subsection{Ablation Study}

To evaluate the contribution of each component in \sysname{}, we construct controlled ablation variants on MAD-Eval (Table~\ref{tab:ablation}). Each variant removes one module while keeping the remaining pipeline unchanged. All experiments use the Qwen3-VL backbone to isolate the contribution of the proposed components.

\noindent\textbf{A1 (VLM-only narration).}
Direct VLM narration without structured fact extraction, identity tracking, verification, or narrative memory.

\noindent\textbf{A2 (No schema).}
Replaces structured facts with scene summaries while preserving the sequential pipeline.

\noindent\textbf{A3 (No memory).}
Removes cross-clip narrative memory while retaining per-clip fact extraction and verification.

\noindent\textbf{A4 (No identity).}
Removes persistent identity tracking, leaving facts in generic form without consistent character grounding.

\noindent\textbf{A5 (No IMDb metadata).}
Disables public movie metadata retrieval while retaining identity tracking, structured facts, verification, and narrative memory.

Table~\ref{tab:ablation} shows that each proposed component contributes to \sysname{}, with every ablation lowering MAD-Eval scores. Removing identity grounding (A4) leads to relative drops of 16.4\% in CIDEr, 16.4\% in SPICE, and 19.6\% in ROUGE-L, even though public movie metadata remains available. This suggests that external metadata alone is insufficient for maintaining narrative consistency, and that persistent character resolution provides complementary information for coherent storytelling. A5 further isolates IMDb metadata: disabling it lowers CIDEr/SPICE/ROUGE-L by 19.6\%/28.4\%/20.3\%, showing that public metadata helps but is not the sole driver of performance.
  
  \begin{table}[t]
\centering
\caption{Ablation results on MAD-Eval. A1--A4 remove one internal
module while keeping the remaining modules unchanged. A5 disables
public IMDb metadata only; identity, schema, and memory remain active.}
\vspace{-0.1in}
\label{tab:ablation}
\small
\setlength{\tabcolsep}{4pt}
\resizebox{\linewidth}{!}{%
\begin{tabular}{lccccccc}
\toprule
Variant & Public Meta. & Identity & Schema & Memory & CIDEr & SPICE & ROUGE-L \\
\midrule
\sysname{} (Full) & \checkmark & $\checkmark$ & $\checkmark$ & $\checkmark$ & \textbf{21.4}& \textbf{6.7}& \textbf{15.3}\\
\midrule
A1 (VLM only) & $\times$ & $\times$ & $\times$ & $\times$ & 11.4 & 4.3 & 9.6 \\
A2 (No schema) & \checkmark & $\checkmark$ & $\times$ & $\checkmark$ &  15.4&  4.4&  12.5\\
A3 (No memory) & \checkmark & $\checkmark$ & $\checkmark$ & $\times$ &  18.0&  6.0&  13.2\\
A4 (No identity) & \checkmark & $\times$ & $\checkmark$ & $\checkmark$ & 17.9 & 5.6 & 12.3 \\
A5 (No IMDb) & $\times$ & $\checkmark$ & $\checkmark$ & $\checkmark$ & 17.2 & 4.8 & 12.2 \\
\bottomrule
\end{tabular}
}
\vspace{-0.1in}
\end{table}

\subsection{\benchname{} Narrative Evaluation}

Table~\ref{tab:storyqa} reports results on the \benchname{} benchmark.
In Track~A, which evaluates understanding within a single video segment,
\sysname{} consistently outperforms AutoAD-Zero across all segment lengths.
The gains remain clear for longer segments, suggesting that narrative-aware descriptions better capture story-level information beyond local clip content.

Track~B evaluates contextual reasoning by generating questions using preceding context while providing only the ADs of the 30-second main clip during evaluation. Across all settings, \sysname{} achieves higher accuracy than AutoAD-Zero, with the largest improvement of +11.5 points in the 90+30 setting. These results indicate that the generated narration more effectively preserves contextual cues such as character identities and ongoing events, enabling the answering model to recover information originating from earlier clips.

The ``Reference AD'' row uses the professional AD text supplied with MAD-Eval as the answerer's input. Because professional AD is scene-limited and may omit details needed for some generated questions, its accuracy is not necessarily 100\%.

Table~\ref{tab:storyqa_ablation} presents component ablations on \benchname{}. Several simplified variants remain competitive on short or local settings, but the complete system achieves the best performance on the longest and most context-dependent settings: 240s in Track~A and 90+30 in Track~B. Removing narrative memory (A3), the identity graph (A4), or IMDb metadata (A5) reduces performance in these long-range settings, highlighting the role of each component. Notably, even without IMDb metadata, A5 remains competitive with or above AutoAD-Zero across all settings, indicating that the gains are not solely attributable to public movie information.

\begin{table*}[t]
\centering
\caption{\textbf{\benchname{} QA benchmark (Accuracy).}
(a) Segment-only QA. (b) Context-conditioned QA with $c$ seconds of prior context plus a 30\,s main clip; only the 30\,s main-clip ADs are provided during evaluation. Reference AD denotes professional human-written AD text.}
\label{tab:storyqa}
\vspace{-0.15in}
\begin{subtable}[t]{0.49\textwidth}
\centering
\caption{Track A: Segment-only QA}
\vspace{-0.1in}
\small
\begin{tabular}{lcccc}
\toprule
Model & 30 & 60 & 120 & 240 \\
\midrule
{Reference AD}& 0.953 & 0.963 & 0.995 & 0.981 \\
\midrule
AutoAD-Zero   & 0.840 & 0.830 & 0.896 & 0.917 \\
\textbf{\sysname{}}     & \textbf{0.892} & \textbf{0.928} & \textbf{0.953} & \textbf{0.972} \\
\bottomrule
\end{tabular}
\end{subtable}
\hfill
\begin{subtable}[t]{0.49\textwidth}
\centering
\caption{Track B: Context-conditioned QA}
\vspace{-0.1in}
\small
\begin{tabular}{lccc}
\toprule
Model & 30+30 & 60+30 & 90+30 \\
\midrule
{Reference AD}& 0.828& 0.825& 0.806\\
\midrule
AutoAD-Zero         & 0.662& 0.654& 0.673\\
\textbf{\sysname{}}       & \textbf{0.754}& \textbf{0.716}& \textbf{0.788}\\
\bottomrule
\end{tabular}
\end{subtable}
\vspace{-0.1in}
\end{table*}

\begin{table*}[t]
\centering
\caption{Component ablations on \benchname{} accuracy. A1: VLM only; A2: no structured schema; A3: no narrative memory; A4: no identity graph; A5: no IMDb metadata.}
\label{tab:storyqa_ablation}
\small
\begingroup
\setlength{\tabcolsep}{8pt}
\begin{tabular}{lccccccc}
\toprule
& \multicolumn{4}{c}{Track A} & \multicolumn{3}{c}{Track B} \\
\cmidrule(lr){2-5}\cmidrule(lr){6-8}
Variant & 30s & 60s & 120s & 240s & 30+30 & 60+30 & 90+30 \\
\midrule
Full
& 0.892 & \textbf{0.928} & \textbf{0.953} & \textbf{0.972}
& 0.754 & 0.716 & \textbf{0.788} \\

A1
& 0.908 & 0.904 & 0.928 & 0.911
& 0.747 & \textbf{0.750} & 0.751 \\

A2
& 0.905 & 0.909 & 0.932 & {0.931}
& \textbf{0.796} & 0.709 & 0.751 \\

A3
& 0.899 & 0.907 & 0.918 & 0.891
& 0.756 & 0.712 & 0.779 \\

A4
& \textbf{0.909} & 0.904 & 0.908 & 0.921
& 0.794 & \textbf{0.750} & 0.774 \\

A5
& 0.892 & 0.909 & 0.908 & 0.901
& 0.761 & 0.716 & 0.770 \\
\bottomrule
\end{tabular}
\endgroup
\end{table*}
\subsection{Qualitative Results}
\begin{figure}[t]
\centering
\includegraphics[width=\linewidth]{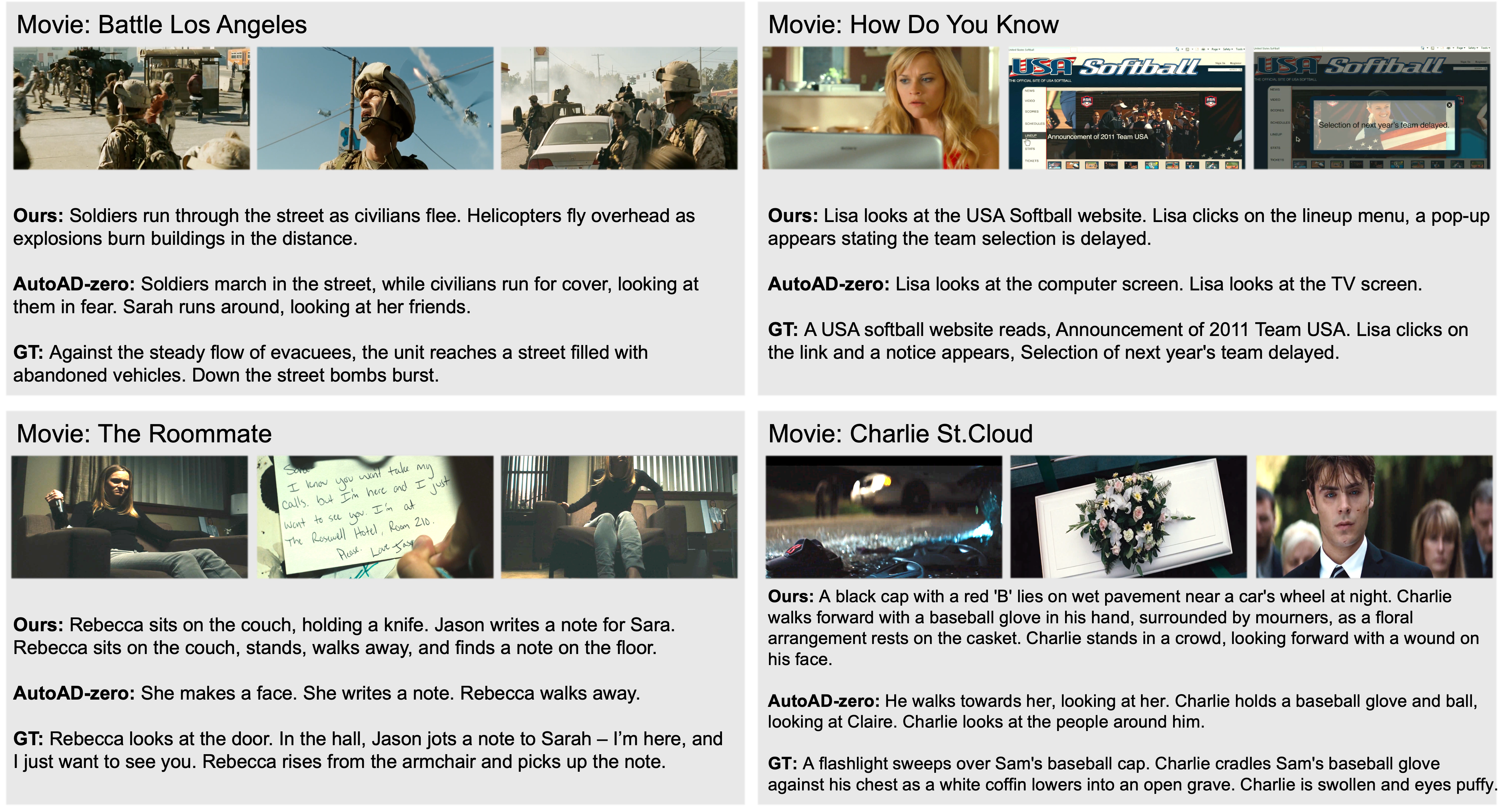}
\caption{\textbf{Qualitative comparison} of \sysname{}, AutoAD-Zero~\cite{autoadzero}, and professional reference AD. (GT).}
\vspace{-0.2 in}
\label{fig:qualitative}
\end{figure}

Figure~\ref{fig:qualitative} compares professional reference AD, AutoAD-Zero, and \sysname{}, illustrating differences in identity consistency and narrative recall across clips. Additional qualitative examples are provided in Supplementary Fig.~S1 and Secs.~S2--S3; human evaluation and preference results are reported in Sec.~S9.

%% file: sec/6_conclusion.tex
\section{Conclusion}

We introduced \sysname{}, a training-free framework for long-form audio description that uses narrative memory to capture evolving story context across scenes. By maintaining an identity graph and structured narrative memory, the system generates coherent descriptions without task-specific training or curated movie resources; public movie metadata can suggest context, but only video-supported facts enter memory. We also introduced \benchname{}, a benchmark for evaluating whether generated AD supports narrative understanding. Experiments show that narrative-aware modeling improves contextual grounding and story comprehension in long-form video description.

\section*{Acknowledgments}
This work was supported by startup funds provided by Dartmouth College. The authors also acknowledge support from the National Science Foundation under CAREER Award No. 2541968.

%% file: Supplementary_Material/supp_src_codes/sec/7_supple.tex
\label{sec/sec7_supple}

\setcounter{page}{1}

\renewcommand{\thefigure}{S\arabic{figure}}
\renewcommand{\thetable}{S\arabic{table}}
\renewcommand{\thesection}{S\arabic{section}}
\renewcommand{\thesubsection}{S\arabic{section}.\arabic{subsection}}
\renewcommand{\theequation}{S\arabic{equation}}

\setcounter{figure}{0}
\setcounter{table}{0}
\setcounter{equation}{0}

\setcounter{section}{0}
\setcounter{subsection}{0}

\section{Overview}

This supplementary document provides additional details and experimental results supporting the main paper. In particular, we include:

\begin{itemize}
\item Additional qualitative comparisons between StoryTeller, AutoAD-Zero, and professional reference AD.
\item Failure cases of baseline methods and discussion of how StoryTeller improves narrative consistency.
\item Analysis of narrative memory dynamics, including visualization of memory weights over time.
\item A backbone comparison explaining how the VideoLLaMA2 and Qwen3-VL configurations differ across MAD-Eval and \benchname{} results.
\item Further details and validation of the StoryAD-QA benchmark.
\item Human evaluation of generated audio descriptions.
\item Implementation details and hyperparameters used in the proposed framework.
\item Full prompt templates used in the StoryTeller pipeline.
\end{itemize}

These materials aim to improve transparency and reproducibility of the proposed framework.

\section{Additional Qualitative Results}
\begin{figure}
    \centering
    \includegraphics[width=1\linewidth]{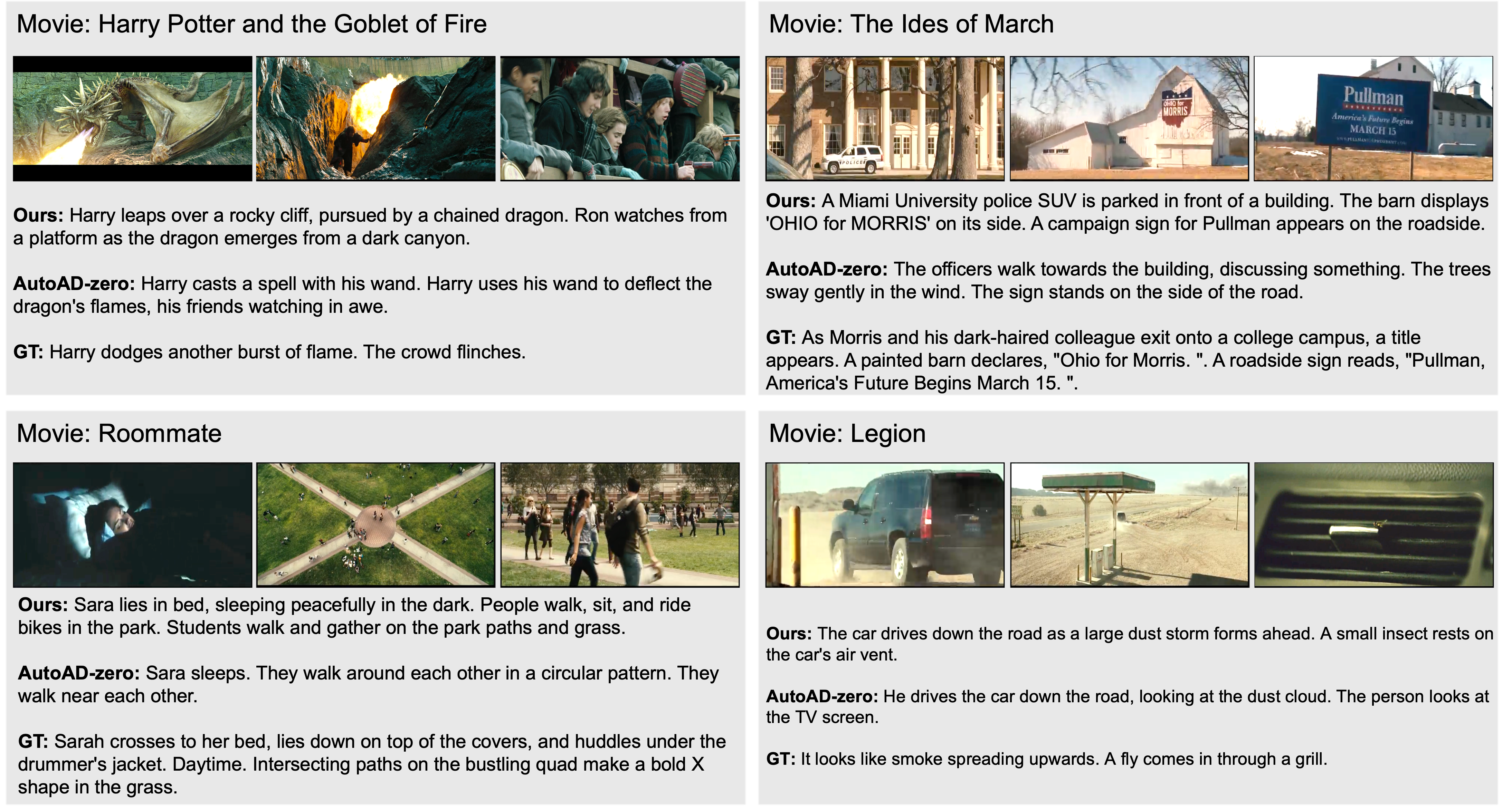}
    \caption{
Additional qualitative comparisons between StoryTeller, AutoAD-Zero,
and professional reference AD across four movies.
StoryTeller captures broader narrative context and scene semantics,
while AutoAD-Zero often focuses on isolated actions or produces
generic descriptions.
}
    \label{fig:supp_quality}
\end{figure}

This section supports the qualitative analysis in Section~5.5 of the main paper by providing more examples beyond the main-paper figure.

Figure~\ref{fig:supp_quality} presents additional qualitative comparisons between StoryTeller, AutoAD-Zero, and professional reference AD across four movies: Harry Potter and the Goblet of Fire, \emph{The Ides of March}, \emph{Roommate}, and \emph{Legion}. Each example shows three representative frames together with the generated audio descriptions.

Across these examples, StoryTeller generally produces descriptions that include more concrete scene context visible in the frames, while AutoAD-Zero often focuses on shorter action-oriented phrases or more generic descriptions. The differences can be observed directly in the
examples.

In \emph{Harry Potter and the Goblet of Fire}, the frames show Harry fighting with the chained dragon. StoryTeller describes Harry leaping over a rocky cliff while being pursued by the dragon, capturing the spatial interaction between the character and the environment. AutoAD-Zero instead focuses on Harry casting a spell with his wand. Both descriptions refer to the same moment, but they emphasize different aspects of the scene.

The example from \emph{The Ides of March} illustrates a scene with visible campaign signage and buildings in the background. StoryTeller describes the police vehicle and the political signs visible in the frames. In contrast, AutoAD-Zero produces a more generic description about officers walking toward a building, which does not explicitly mention the visual text present in the scene.

In the \emph{Roommate} example, the first frame shows a character lying in bed, followed by outdoor campus scenes. StoryTeller describes both the indoor and outdoor settings shown across the frames. AutoAD-Zero produces a shorter description focused mainly on the sleeping character and nearby people walking.

Finally, the \emph{Legion} example shows a car driving along a desert road while a dust storm forms in the distance. StoryTeller describes the car moving toward the storm and also mentions a small insect visible on the car vent in the final frame. AutoAD-Zero instead produces a more general description of the car driving and a person looking toward a
dust cloud.

These examples highlight how the generated descriptions differ in content emphasis across methods. StoryTeller tends to include environmental details visible in the frames, while AutoAD-Zero often produces shorter descriptions centered on basic actions.

\section{Comparative Error Analysis}
\begin{figure}
    \centering
    \includegraphics[width=1\linewidth]{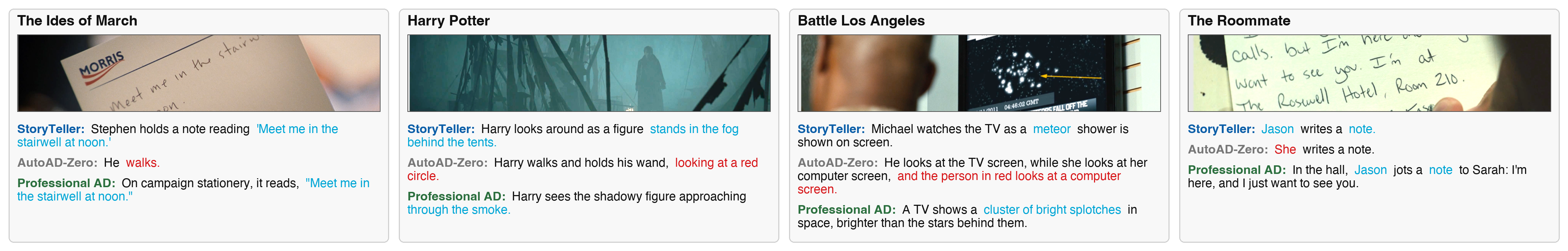}
    \caption{\textbf{Qualitative comparison of audio descriptions.} Examples comparing StoryTeller, AutoAD-Zero, and professional reference AD across several movies. Text highlighted in \textcolor{cyan}{blue} corresponds to information that is visually grounded in the frames and appears in both StoryTeller and the professional reference AD. Text highlighted in \textcolor{red}{red} marks content that is less consistent with the visual evidence in the scene. These examples illustrate common challenges in automatic audio description, including grounding textual cues, interpreting scene entities, and maintaining character identity across scenes.}
    \label{fig:supp_failures}
\end{figure}

Figure~\ref{fig:supp_failures} presents representative qualitative examples comparing StoryTeller, AutoAD-Zero~\cite{autoadzero}, and professional reference AD across several movies. These examples highlight common challenges in automatic audio description, including difficulties in grounding textual cues, identifying scene entities, and maintaining character identity across scenes.

In the figure, text highlighted in \textcolor{cyan}{blue} corresponds to information that is consistent with the visual frames and appears in both StoryTeller and the professional reference AD. Text highlighted in \textcolor{red}{red} marks content that is less consistent with the visual evidence in the scene.

\noindent\textbf{Missing textual or symbolic cues.}
In the \textit{The Ides of March} example, the frame clearly shows a note containing the message “Meet me in the stairwell at noon.” StoryTeller and the professional reference AD both capture this textual information, while the baseline description instead focuses on a generic action (“He walks”), overlooking the written content visible in the scene.

\noindent\textbf{Object grounding ambiguity.}
In the \textit{Harry Potter and the Goblet of Fire} example, the scene shows a figure approaching through fog behind the tents. StoryTeller and the professional reference AD describe the approaching figure, whereas the baseline description refers to a “red circle,” which does not clearly correspond to any object visible in the scene.

\noindent\textbf{Scene interpretation differences.}
The \textit{Battle: Los Angeles} example illustrates a case where the baseline description interprets the television display as people interacting with computer screens. In contrast, StoryTeller and the professional reference AD identify the meteor shower shown on the display.

\noindent\textbf{Character identity ambiguity.}
In the final example, the scene shows a character writing a note. StoryTeller correctly identifies the character as Jason and describes the action accordingly, while the baseline description refers to the character as “she,” which loses the identity information present in the narrative context.

These examples highlight common challenges in grounding audio descriptions to visual evidence and maintaining narrative consistency across scenes. By incorporating structured narrative facts and an evolving memory mechanism, StoryTeller is able to produce descriptions that better preserve scene context and character identity.

\section{Implementation Details}

\subsection{Model Configuration}
We use the following models in the StoryTeller pipeline:

\begin{itemize}[leftmargin=1.5em]
    \item \textbf{VLM for scene summarization and fact extraction:} Qwen3-VL-2B-Instruct 
    \item \textbf{VLM for fact verification:} Qwen3-VL-8B-Thinking
    \item \textbf{Text embedding model:} Qwen3-VL-Embedding-2B
    \item  \textbf{VLM for StoryAD-QA generation and answering:} Gemini3-Flash for initial question generation and answering; Gemini 3.5 Flash is used only to regenerate questions that fail the first manual verification pass.
    \item \textbf{Face recognition encoder:} ArcFace \cite{deng2019arcface}
\end{itemize}

Within each reported configuration, the same generation backbone is used consistently unless a table explicitly compares backbones. Scene-level observations are first summarized into structured facts, which are then verified and stored in the narrative memory. During generation, StoryTeller retrieves the most relevant verified facts and identity-consistent context to produce the final audio description. The AD pipeline may also retrieve IMDb plot summaries, character lists, and short character descriptions using the movie title. These texts can suggest possible names or plot context, but are not copied into narration or treated as professional reference AD; accepted facts must be supported by the clip.

Except for Gemini, all models are executed locally on a cluster comprising 8 {NVIDIA A100-SXM4-80GB} GPUs and 8 {NVIDIA RTX 6000 Ada Generation} GPUs. On one allocated A100-SXM4-80GB GPU, generating audio descriptions for a single film requires approximately 3--4 hours, corresponding to roughly 650 clips per movie. Across the 10 full movie runs, the pipeline processes 6,520 clips and 13,974 candidate facts; verification takes $3.68\pm1.73$ seconds per fact block, uses 473K tokens in total, and accepts 94.7\% of candidate facts. The coarse semantic gate is approximately $3{,}000\times$ faster than VLM verification. Generating \benchname{} questions for the 10-movie evaluation set takes approximately 10--15 minutes per track.

In practice, generating the complete StoryAD-QA benchmark (2,574 questions across 10 movies) incurs an API cost of under \$20, while evaluating the complete benchmark costs less than \$5. These values are approximate and may vary depending on API pricing and the model's internal reasoning-token usage.

For \benchname{} question generation only, Gemini3-Flash receives the video and a short Wikipedia movie overview. During question answering, it receives only the generated AD text, question, and answer choices---never the source video, movie title, IMDb or Wikipedia text, subtitles, or other retrieved text.

\subsection{Hyperparameters}
\label{sec:supp_hyperparameters}

Table~\ref{tab:hyperparams} summarizes the hyperparameters used in StoryTeller. These values are fixed globally across all experiments and are not tuned on the evaluation datasets.

\begin{table}[t]
\caption{Hyperparameters used in StoryTeller.}
\centering
\begin{tabular}{l c}
\hline
Parameter & Value \\
\hline
Identity threshold $\tau$ & 0.58 \\
Grounding threshold $\sigma$ & 0.20 \\
Reinforcement strength $\alpha$ & 0.35 \\
Decay factor $\lambda$ & 0.92 \\
Initial memory weight $w_m^{(0)}$ & 0.25 \\
\hline
\end{tabular}
\label{tab:hyperparams}
\end{table}

\noindent\textbf{Grounding Threshold ($\sigma$).}
The semantic grounding threshold determines which candidate facts are forwarded to the computationally expensive VLM verification stage. To calibrate this threshold, we executed the complete StoryTeller pipeline on a separate set of movies and recorded the grounding score of every extracted candidate fact, producing approximately 2,700 candidate facts.

Among candidate facts ultimately accepted by the verifier, the minimum observed similarity was approximately 0.21, the 5th percentile was 0.36, and the median was 0.55. Based on these statistics, we selected $\sigma=0.20$, slightly below the minimum observed similarity, so that the semantic filter functions as a high-recall gate. Its purpose is not to decide factual correctness, but to remove clearly unrelated candidates before VLM verification.
\begin{table}[t]
\centering
\caption{Effect of different semantic grounding thresholds. ``Verifier calls saved'' is measured relative to $\sigma=0.20$.}
\label{tab:threshold_sweep}
\begin{tabular}{c@{\hspace{0.9cm}}c@{\hspace{1.0cm}}c}
\toprule
$\sigma$ & Calls saved & Accepted facts \\
\midrule
0.20 & 0.0\% & 100.0\% \\
0.40 & 1.4\% & 98.7\% \\
0.45 & 6.1\% & 94.3\% \\
0.50 & 19.3\% & 81.6\% \\
0.55 & 43.2\% & 57.9\% \\
\bottomrule
\end{tabular}
\end{table}

Table~\ref{tab:threshold_sweep} reports the trade-off between computational cost and recall under alternative threshold values. Higher thresholds reduce the number of verifier calls but also discard visually supported facts. We therefore adopt the conservative setting $\sigma=0.20$ to prioritize narrative recall over computational savings.

\noindent\textbf{Identity Threshold ($\tau$).}
Identity assignment uses a fixed cosine similarity threshold of $\tau=0.58$ for ArcFace embeddings. This value was selected empirically to balance missed associations and incorrect identity merges, and is kept fixed across all experiments.

\noindent\textbf{Hyperparameter Robustness.}
Although \sysname{} introduces several global hyperparameters (identity threshold $\tau$, grounding threshold $\sigma$, reinforcement strength $\alpha$, and decay rate $\lambda$), these are fixed once and not tuned on evaluation data. To assess robustness, we vary $\alpha$ within $\pm20\%$ and $\lambda$ within $\pm5\%$ while keeping the remaining parameters fixed. Performance on the test split remains stable, with StoryAD-QA accuracy varying by less than 3 percentage points and CIDEr by less than 6 percentage points. Results are summarized in Table~\ref{tab:robustness}.

\begin{table}[t]
\centering
\caption{Hyperparameter robustness on the test split. Performance remains stable under moderate variation. \benchname{} results are reported as accuracy (\%) for Track~A. The best performance for each metric is bolded.}
\label{tab:robustness}
\setlength{\tabcolsep}{8pt}

\begin{tabular}{l c cccc}
\toprule
\textbf{Setting} & \textbf{CIDEr} & \multicolumn{4}{c}{\textbf{\benchname{} Accuracy (\%)}} \\
\cmidrule(lr){3-6}
 &  & \textbf{30s} & \textbf{60s} & \textbf{120s} & \textbf{240s} \\
\midrule

Default
& 21.4
& 89.2 & 92.8 & 95.3 & 97.2 \\

\midrule

$\alpha$ -20\%
& 20.2
& 89.4 & \textbf{94.4} & 96.2 & 97.2 \\

$\alpha$ +20\%
& 21.0
& \textbf{90.5} & \textbf{94.4} & \textbf{97.2} & \textbf{98.2} \\

\midrule

$\lambda$ -5\%
& \textbf{21.9}
& 89.7 & 94.2 & 95.3 & 97.2 \\

$\lambda$ +5\%
& 21.8
& 90.1 & 94.0 & 94.8 & 96.3 \\

\bottomrule
\end{tabular}
\end{table}

\subsection{Memory Overhead of Identity Reseeding}

Identity reseeding maintains a small backlog of unresolved (anonymous) tracklets until sufficient evidence is available for identity assignment. Each unresolved tracklet stores a single 512-dimensional ArcFace embedding (approximately 2\,KB). Across all movies in our experiments, retaining this anonymous backlog required less than 3.5\,MB per movie, indicating that delayed identity assignment introduces negligible memory overhead.

\subsection{Limitations and Discussion}

StoryTeller maintains consistent character references using the identity graph described in Section~3.1 of the main paper which associates face tracklets across clips through clustering and similarity-based assignment. In rare cases, clustering errors may temporarily associate a tracklet with the wrong identity. When this occurs, the generated description may mention an incorrect character name for the corresponding scene. 

Because StoryTeller maintains a persistent narrative state, such errors can often be corrected when stronger visual or narrative evidence becomes available. In particular, the reseeding mechanism described in Section~3.1 of the main paper allows identities to be updated once additional observations clarify the character assignment. However, if incorrect associations persist across several clips, the resulting descriptions may temporarily reduce narrative accuracy. 

Interestingly, this behavior also highlights the importance of maintaining an explicit narrative memory rather than relying solely on frame-level captioning. By storing and updating structured narrative facts over time, the system can revise earlier assumptions and maintain consistent story context as additional evidence appears. Future work may further improve robustness by integrating stronger identity verification or multimodal reasoning mechanisms.

\section{Analysis of Narrative Memory Dynamics}

To better understand how StoryTeller maintains narrative context across long video sequences, we analyze the evolution of memory weights associated with verified narrative facts. Each fact $f_m$ stored in the narrative memory is associated with a salience weight
$w_m^{(t)}$ that reflects its current narrative importance. When later scenes
provide visual or semantic evidence supporting $f_m$, the memory module
reinforces its salience by increasing $w_m^{(t)}$. Conversely, if $f_m$ is not
observed again, its salience gradually decreases through decay.

Figure~\ref{fig:memory_curves} visualizes representative trajectories from three movies: \emph{Signs}, \emph{Battle: Los Angeles}, and \emph{Harry Potter and the Goblet of Fire}. For each movie we plot two facts: a persistent narrative cue (solid line) and a transient observation (dashed line). These trajectories illustrate how StoryTeller selectively retains story-relevant information while allowing short-lived visual details to fade from memory.

\subsection{Memory Weight Evolution}
\begin{figure*}[tb]
\centering
\begin{subfigure}{0.33\linewidth}
    \centering
    \includegraphics[width=\linewidth]{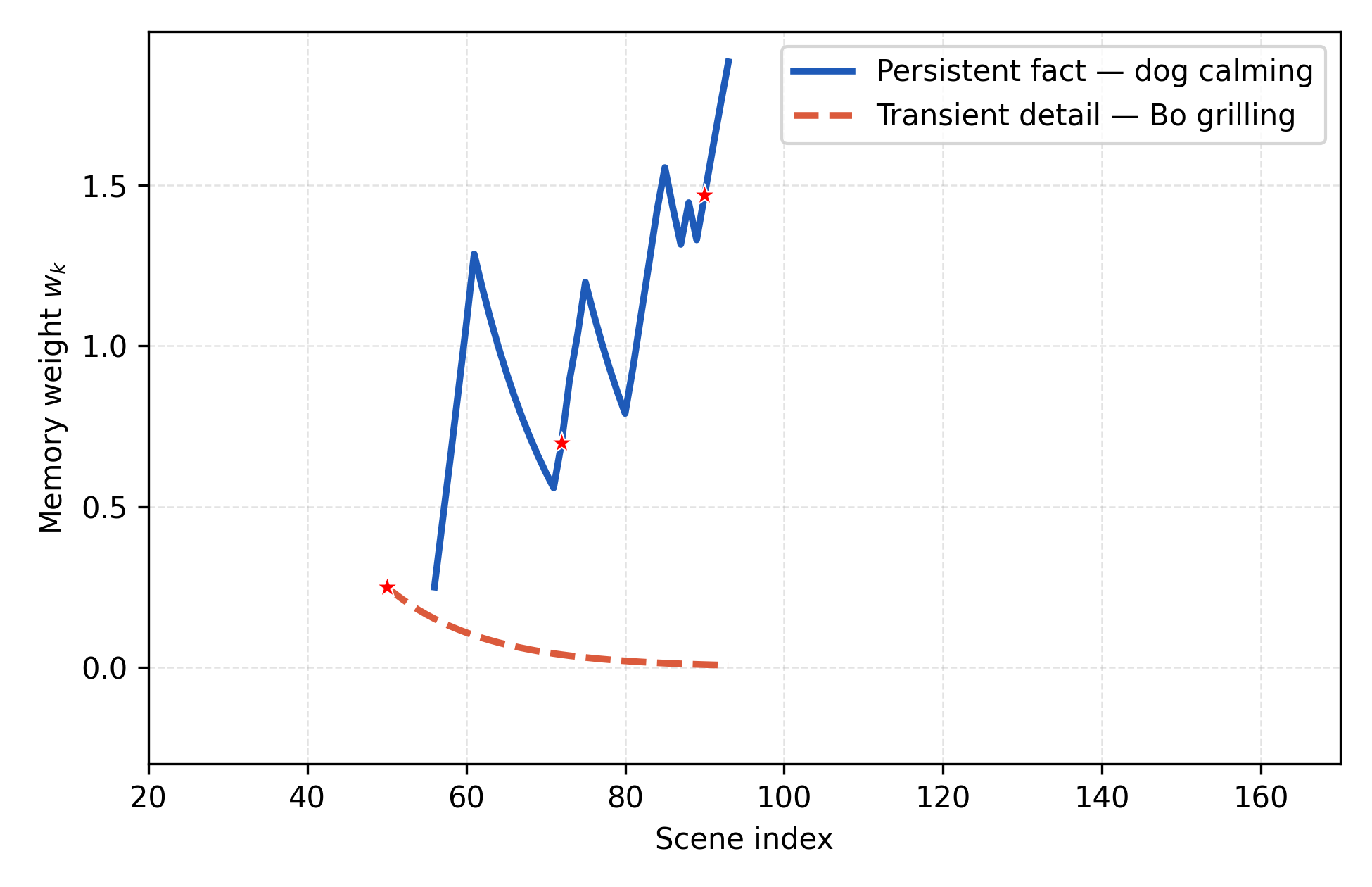}
    \caption{Signs}
\end{subfigure}\hfill
\begin{subfigure}{0.33\linewidth}
    \centering
    \includegraphics[width=\linewidth]{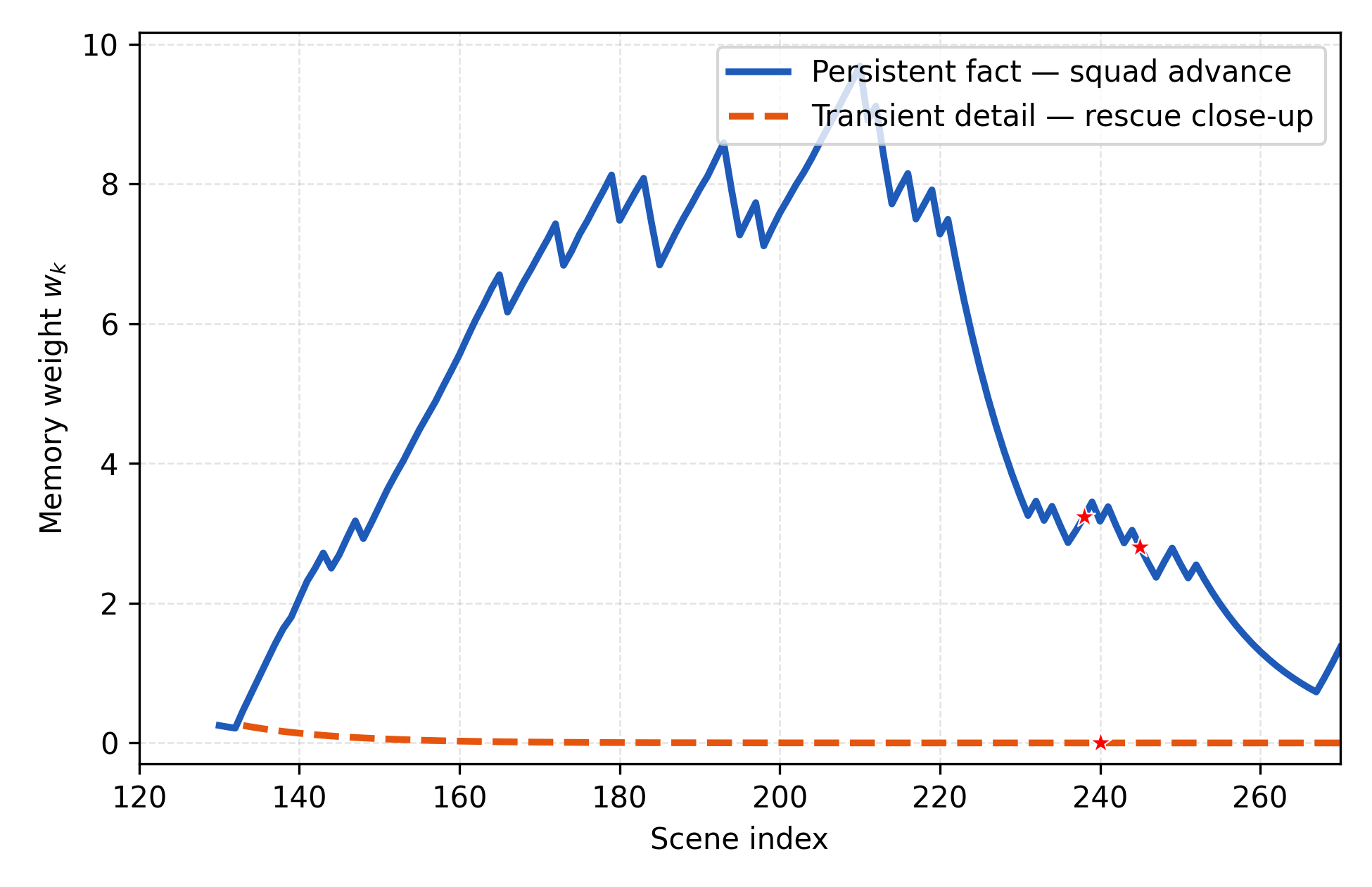}
    \caption{Battle: Los Angeles}
\end{subfigure}\hfill
\begin{subfigure}{0.33\linewidth}
    \centering
    \includegraphics[width=\linewidth]{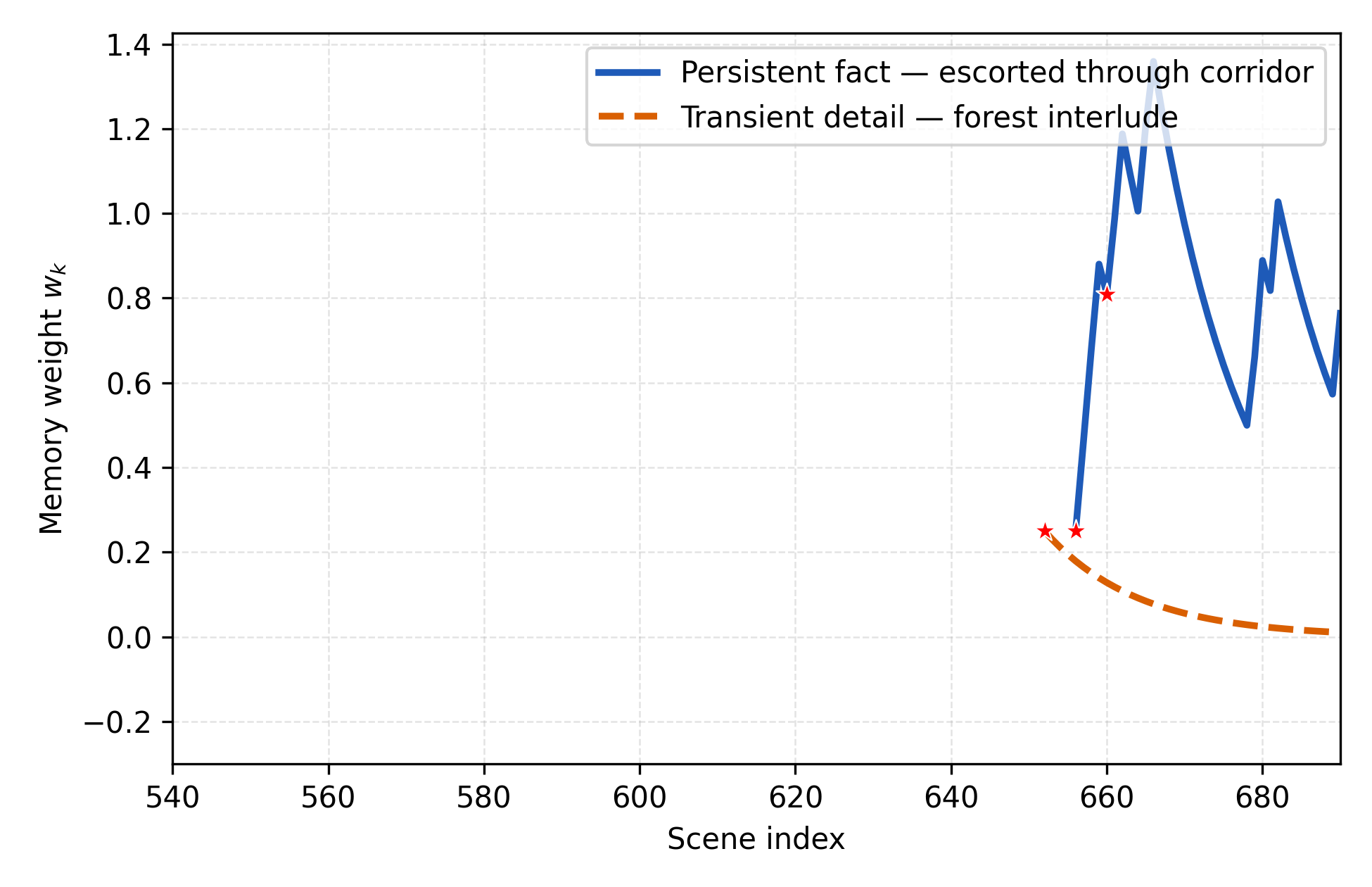}
    \caption{Goblet of Fire}
\end{subfigure}

\caption{
Memory-weight trajectories $w_m^{(t)}$ for representative facts $f_m$ extracted from
\emph{Signs}, \emph{Battle: Los Angeles}, and
\emph{Harry Potter and the Goblet of Fire}.
Solid curves denote \textbf{persistent narrative cues} that reappear across
multiple scenes and therefore receive repeated reinforcement by the
memory update mechanism.
Dashed curves denote \textbf{transient observations} that occur only once
and gradually decay when they are not reinforced.
Star markers indicate the scene indices used in the qualitative
case studies shown in Fig.~\ref{fig:memory_frames}, where the corresponding
video frames and StoryTeller-generated audio descriptions are displayed.
}
\label{fig:memory_curves}
\end{figure*}

\begin{figure}[h]
    \centering
    \includegraphics[width=0.8\linewidth]{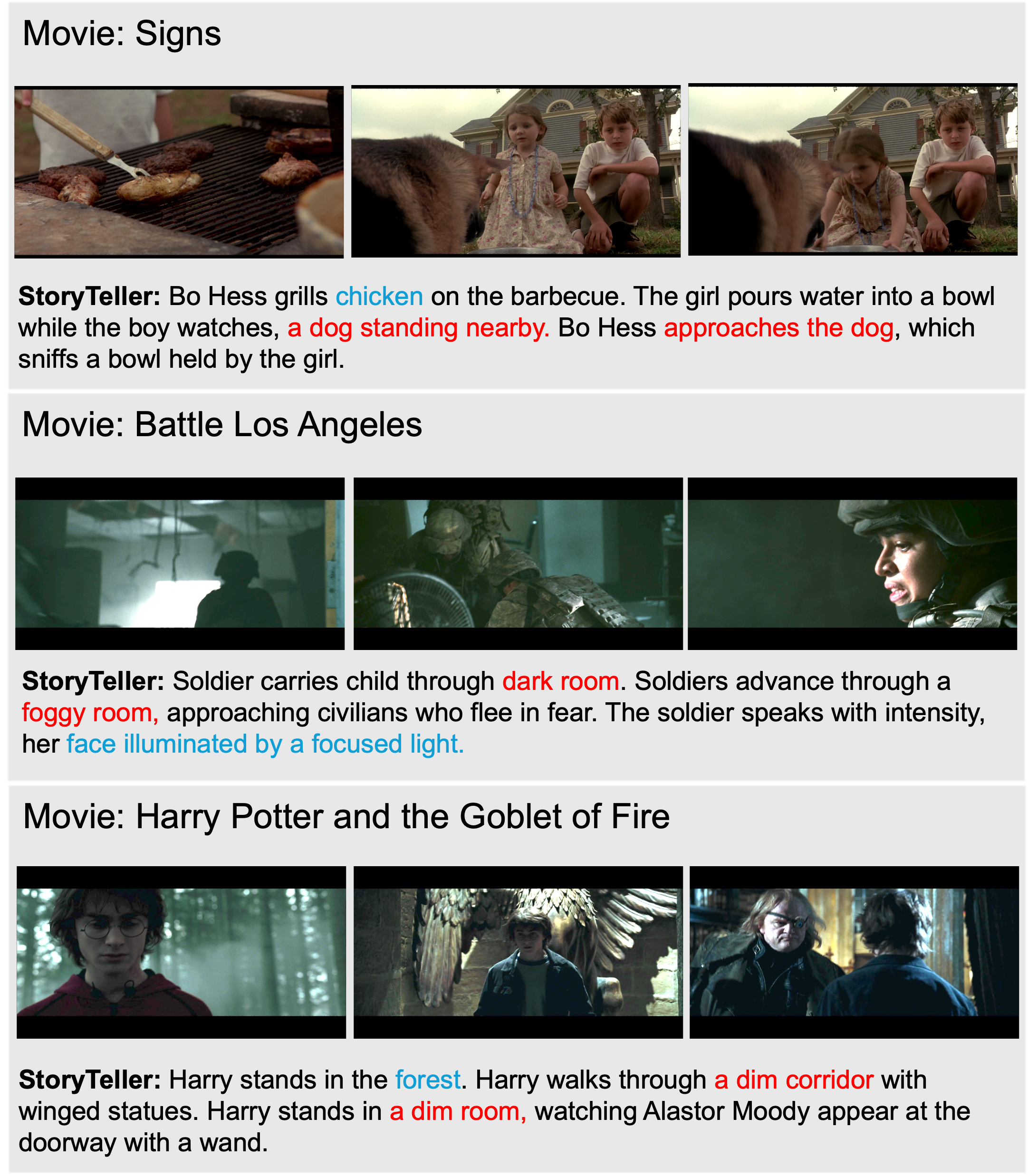}
    \caption{
Representative frames corresponding to the facts highlighted in
Fig.~\ref{fig:memory_curves}, together with StoryTeller-generated
audio descriptions.
Persistent narrative cues (highlighted in red) remain relevant across
multiple scenes and are reinforced by the memory mechanism.
Transient observations (highlighted in blue) appear only briefly and
quickly decay.
}
\label{fig:memory_frames}
\end{figure}

Figure~\ref{fig:memory_curves} highlights two characteristic patterns of the memory update mechanism. Solid curves correspond to \textbf{persistent narrative facts} whose weights increase when similar visual evidence appears in later scenes. Dashed curves correspond to \textbf{transient observations} that occur only once and therefore decay when they are not reinforced. Star markers indicate scene indices selected for the qualitative examples shown in Fig.~\ref{fig:memory_frames}.

\paragraph{Signs.}
In \emph{Signs}, the persistent fact corresponds to the interaction between the children and the family dog near the farmhouse porch. Because similar configurations recur across several scenes, the memory module repeatedly reinforces this fact, resulting in a steadily increasing weight. In contrast, a transient observation involving Bo grilling food outside appears only briefly and is never revisited, causing the associated weight to decay rapidly.

\paragraph{Battle: Los Angeles.}
For \emph{Battle: Los Angeles}, the persistent fact captures the squad advancing through a dark, smoke-filled environment while protecting
civilians. Multiple neighboring scenes depict variations of this tactical situation, leading to repeated reinforcement and a sustained memory weight. By comparison, a brief close-up of a wounded soldier appears only once during the moment and quickly fades from
memory.

\paragraph{Harry Potter and the Goblet of Fire.}
The third example contrasts a brief forest insert with a repeated corridor sequence. The persistent fact corresponds to Harry being escorted through a stone corridor by Moody. Each return to the hallway configuration reinforces the stored fact and increases its weight. The forest insert, however, functions only as a short atmospheric transition and therefore decays quickly.

\subsection{Case Study}
While the weight trajectories reveal how narrative facts evolve in memory, the qualitative examples in Fig.~\ref{fig:memory_frames} illustrate how this salience affects the generated audio descriptions.

Each panel presents representative frames together with the narration generated by \sysname{}. Words highlighted in red correspond to persistent narrative cues whose memory weights remain high across multiple scenes. Because these facts remain salient, \sysname{} continues referencing the same narrative context even when the camera framing shifts or visual details change.

For example, in \emph{Signs}, the generated description consistently refers to the interaction between the children and the dog despite changes in viewpoint. Similarly, in \emph{Battle: Los Angeles}, the narration preserves the context of soldiers advancing through smoke-filled interiors while protecting civilians. In \emph{Harry Potter and the Goblet of Fire}, repeated corridor shots reinforce the interaction between Harry and Moody as they move through the dim stone corridor.

In contrast, words highlighted in blue correspond to transient visual details that appear only briefly. Because these observations are not reinforced by later scenes, their memory weights decay and they do not influence subsequent descriptions.

These examples demonstrate the intended behavior of the StoryTeller memory mechanism: facts tied to ongoing narrative context accumulate weight and remain retrievable across scenes, while isolated observations gradually fade. This selective retention enables the system to maintain story-level continuity without overloading the memory with short-lived details.

\section{Backbone Model Comparison}

We compare StoryTeller with two generation backbones to separate architectural effects from backbone choice. The VideoLLaMA2 row corresponds to the \benchname{} setting reported in the main narrative-evaluation table, while the Qwen3-VL row corresponds to the default implementation used for the Qwen3-VL MAD-Eval result in the main comparison table.
\begin{table}[h]
\caption{Comparison of different VLM backbones. \benchname{} results are reported as accuracy (\%) for Track~A (blue shade) and Track~B (orange shade).}
\setlength{\tabcolsep}{5pt}
\centering
{%
\begin{tabular}{l c
>{\columncolor{blue!10}}c
>{\columncolor{blue!10}}c
>{\columncolor{blue!10}}c
>{\columncolor{blue!10}}c
>{\columncolor{orange!15}}c
>{\columncolor{orange!15}}c
>{\columncolor{orange!15}}c}
\toprule
\textbf{Backbone} & \textbf{CIDEr} &
\multicolumn{7}{c}{\textbf{StoryAD-QA Accuracy (\%)}} \\
\cmidrule(lr){3-9}
 & & \textbf{30s} & \textbf{60s} & \textbf{120s} & \textbf{240s}
 & \textbf{30+30} & \textbf{60+30} & \textbf{90+30} \\
\midrule
Qwen3-VL
& \textbf{21.4}
& \textbf{89.2} & \textbf{92.8} & \textbf{95.3} & \textbf{97.2}
& \textbf{75.4} & \textbf{71.6} & \textbf{78.8} \\

VideoLLaMA2
& 19.1
& 85.9 & 90.2 & 92.9 & 95.4
& 72.7 & 70.9 & 72.4 \\
\bottomrule
\end{tabular}
}
\end{table}

\section{{Additional Evaluation on MovieChat}}
\label{sec:moviechat_sanity}
We additionally evaluate \sysname{} on a subset of the MovieChat-1K test set~\cite{song2024moviechat}, containing 9 videos and 27 questions. Table~\ref{tab:moviechat} compares methods under the input modality appropriate to each setting. MovieChat follows the standard MovieChat setup and accordingly receives the original source video (visual modality) as input. In contrast, AutoAD-Zero and \sysname{} are AD-generation systems, so we evaluate them by replacing the video with their generated AD text and asking the same LLM answerer to answer the MovieChat questions from text alone. This measures how much long-video QA information is preserved in the generated descriptions.

For a fair no-curated-resource comparison between AD systems, AutoAD-Zero is evaluated without its character bank, and \sysname{} is evaluated without public IMDb metadata. We also report \sysname{} with IMDb metadata to measure the additional benefit of public movie context. As shown in Table~\ref{tab:moviechat}, \sysname{} outperforms AutoAD-Zero under the same generated-AD input setting, remains strong without IMDb metadata, and improves further when public metadata is available. \benchname{} remains our primary narrative benchmark because it provides denser coverage of story events, with approximately one question per 30s compared with three questions per movie in this MovieChat subset.

\begin{table}[t]
\centering
\caption{{\textbf{Additional evaluation on MovieChat} The subset contains 9 videos and 27 questions. 
The subset contains 9 videos and 27 questions. MovieChat$^\dagger$ uses the original source video, while AutoAD-Zero and \sysname{} use generated AD text as input to the same LLM answerer.}}
\label{tab:moviechat}
\small
\begingroup
\setlength{\tabcolsep}{8pt}
\begin{tabular}{llcc}
\toprule
Method & Input & Acc. & Score \\
\midrule
MovieChat & Source video & 0.623 & 3.230 \\
Human reference caption & Short caption & 0.037 & 0.185 \\
AutoAD-Zero (no character bank) & Generated AD & 0.704 & 3.593 \\
\sysname{} (no IMDb) & Generated AD & 0.741 & 3.778 \\
\sysname{} (with IMDb) & Generated AD & \textbf{0.778} & \textbf{3.852} \\
\bottomrule
\end{tabular}
\endgroup
\end{table}

\section{StoryAD-QA Benchmark Overview and Validation}
\label{appx:storyadqa}
This section provides additional details about the construction and validation of the \benchname\ benchmark introduced in Section~4 of the main paper. We summarize the distribution of question types and segment durations, describe the manual verification protocol used for all retained questions, and report validation rates and annotation consistency.

\subsection{Benchmark Overview and Statistics}
Table~\ref{tab:storyad_stats} summarizes the distribution of the final retained \benchname{} questions across the two evaluation tracks. Track~A evaluates models using only the target segment, while Track~B requires additional narrative context from preceding segments.

To capture events unfolding across multiple shots, we construct segments by grouping consecutive clips into temporal windows of varying duration. In Track~A, questions are associated with segments ranging from 30 to 240 seconds. In Track~B, each question is paired with a 30-second target segment and an additional context window preceding it, encouraging models to reason over narrative dependencies across segments.

\begin{table}[t]
\caption{{Distribution of the 2,574 questions in \benchname\ across evaluation tracks and segment durations. Track~A uses only the target segment, while Track~B provides a 30, 60, or 90-second preceding context window plus a fixed 30-second target segment.}}
\centering
\small
\begin{tabular}{lccc}
\toprule
\textbf{Track} & \textbf{Segment Duration} & \textbf{\# Questions} \\
\midrule
Track A (segment-only QA) 
& 30s  & 860 \\
& 60s  & 430 \\
& 120s & 212 \\
& 240s & 109 \\
\midrule
\textbf{Track A Total} & & \textbf{1,611} \\
\midrule
Track B (context-conditioned QA) 
& 30s context + 30s  & 449 \\
& 60s context + 30s  & 294 \\
& 90s context + 30s  & 220 \\
\midrule
\textbf{Track B Total} & & \textbf{963} \\
\bottomrule
\end{tabular}
\label{tab:storyad_stats}
\end{table}

\noindent\textbf{Answer Randomization.}
To prevent positional bias from the model generating the questions, the answer options are randomly shuffled before being provided to the language model during evaluation.

\noindent\textbf{API Usage.} Although the API safety filters were set to the least restrictive available configuration, a small number of AD inputs were still blocked by the service and could not be processed. These instances were treated as missing values and excluded from aggregate results.

\subsection{Manual Verification}

\begin{figure}[t]
\centering
\includegraphics[width=0.9\linewidth]{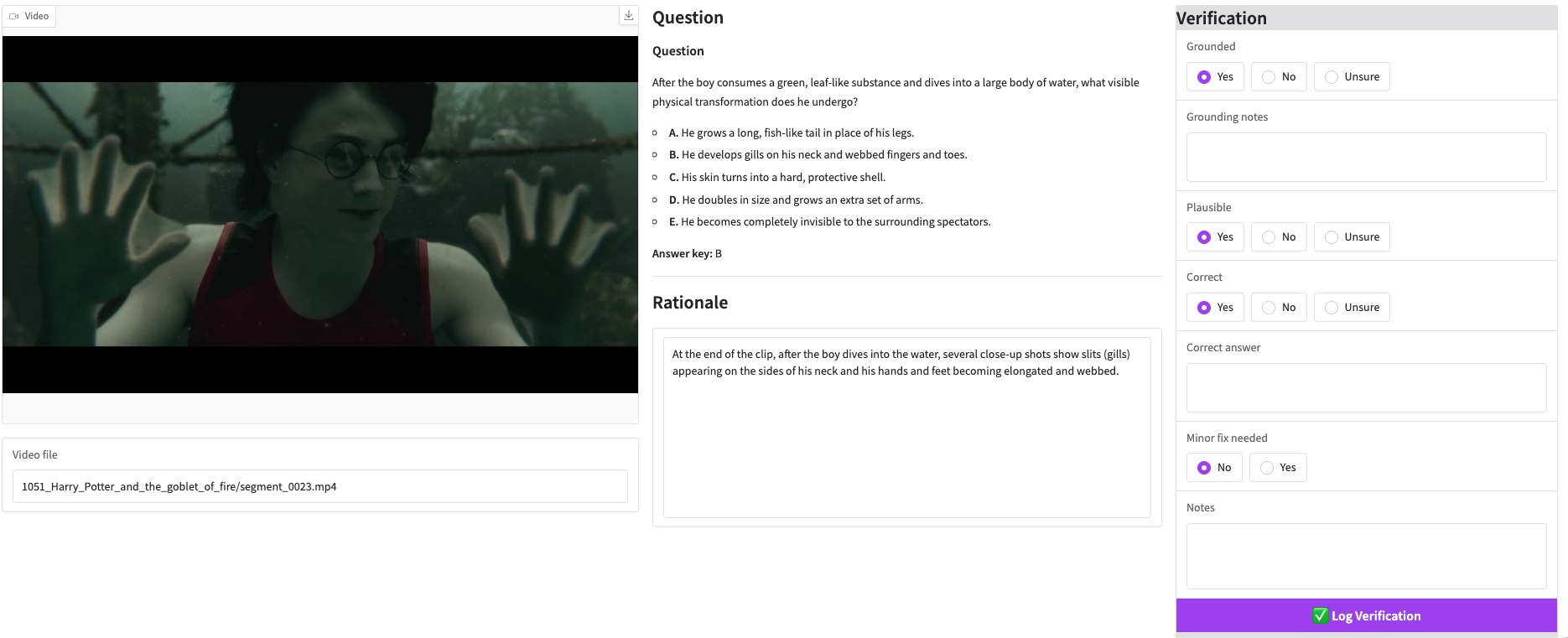}
\caption{
Annotation interface used for human evaluation. Annotators watch the video segment, review the multiple-choice question and answer options, and record judgments for several verification criteria. 
}
\label{fig:s7_interface}
\end{figure}

All questions retained in \benchname{} are manually verified before inclusion. Each question is reviewed through a custom  \textit{Gradio} annotation interface, which presents annotators with the corresponding video segment together with the generated question, answer options, the labeled correct answer, and the accompanying rationale produced during question generation. Figure~\ref{fig:s7_interface} demonstrates the interface from where annotators watch the corresponding video segment and the corresponding generated questions and submit their evaluation of  several quality metrics designed to assess the integrity of the question–answer pairs. The review covers both the question and its answer set: annotators check that the question is visually grounded and unambiguous, that the labeled answer is uniquely supported by the video evidence, and that the distractor options are plausible but incorrect. Questions that fail verification are discarded or regenerated and then checked again before they can enter the final benchmark.

\subsubsection{Verification criteria}

We use \textbf{Correct answer} to assess whether the labeled answer is correct and uniquely supported by the visual evidence in the clip. In addition, we adopt two quality metrics from~\cite{kala2025adqa}. \textbf{Visually grounded} evaluates whether the correct answer can be determined solely from the visual content of the clip, without relying on dialogue, audio, or external knowledge. \textbf{Plausible distractors} assesses whether the incorrect answer options constitute realistic alternatives that could reasonably be confused with the correct answer given the scene.

In addition to the binary \textit{Yes}/\textit{No} labels, annotators could also select \textit{Unsure} when the judgment was genuinely ambiguous. The verification statistics for the full benchmark are presented in Table~\cref{tab:qa_human_eval_full}. 

\begin{table}[t]
\caption{{Human verification results on all questions of the initial \benchname{} benchmark questions. Rates report the proportion of questions satisfying each criterion.}}
\centering
\small
\setlength{\tabcolsep}{8pt} 
\begin{tabular}{l|c|c|c}

\toprule

\textbf{Setting} & \textbf{Grounded} & \textbf{Plausible Distractors} & \textbf{Correct Ans.} \\

\midrule

\multicolumn{4}{l}{\textit{Track A}} \\

30\,s  & 0.890 & 0.857 & 0.752 \\

60\,s  & 0.901 & 0.875 & 0.795 \\

120\,s & 0.892 & 0.887 & 0.814 \\

240\,s & 0.963 & 0.936 & 0.833 \\

\midrule

\multicolumn{4}{l}{\textit{Track B}} \\

60\,s  & 0.838 & 0.827 & 0.744 \\

90\,s  & 0.822 & 0.784 & 0.703 \\

120\,s & 0.886 & 0.831 & 0.699 \\

\midrule

Unsure & 3 & 3 & 45 \\

\bottomrule

\end{tabular}

\label{tab:qa_human_eval_full}
\end{table}

\subsection{Question Regeneration}
Questions that received a negative rating in any evaluation criterion during the initial human verification were discarded and regenerated using the same video clip and prompt, but with a stronger VLM, namely Gemini 3.5 Flash. In total, we regenerated 773 questions, of which 618 were retained after filtering out clips that the VLM determined lacked sufficient information to support a meaningful question. All regenerated questions then underwent a second round of manual verification by a human annotator. Every question in this pass was accepted and included in the final benchmark.

\section{Human Evaluation of Audio Descriptions}

To evaluate the perceptual quality of generated audio descriptions, we conducted a human study comparing StoryTeller with AutoAD-Zero.

\subsection{Study Setup}

We randomly sampled 50 clips from the evaluation set, each with an average duration of approximately 20 seconds. For each clip, three annotators watched the video and compared two candidate audio descriptions: one generated by StoryTeller and one generated by AutoAD-Zero. The descriptions were presented with anonymized labels (\emph{Option A} and \emph{Option B}), and their order was randomized for each trial
to avoid positional bias. Because each clip spans approximately 20 seconds, the candidate descriptions are formed by concatenating multiple consecutive AD sentences generated for that clip, resulting in short multi-sentence descriptions.

\begin{figure}[t]
\centering
\includegraphics[width=\linewidth]{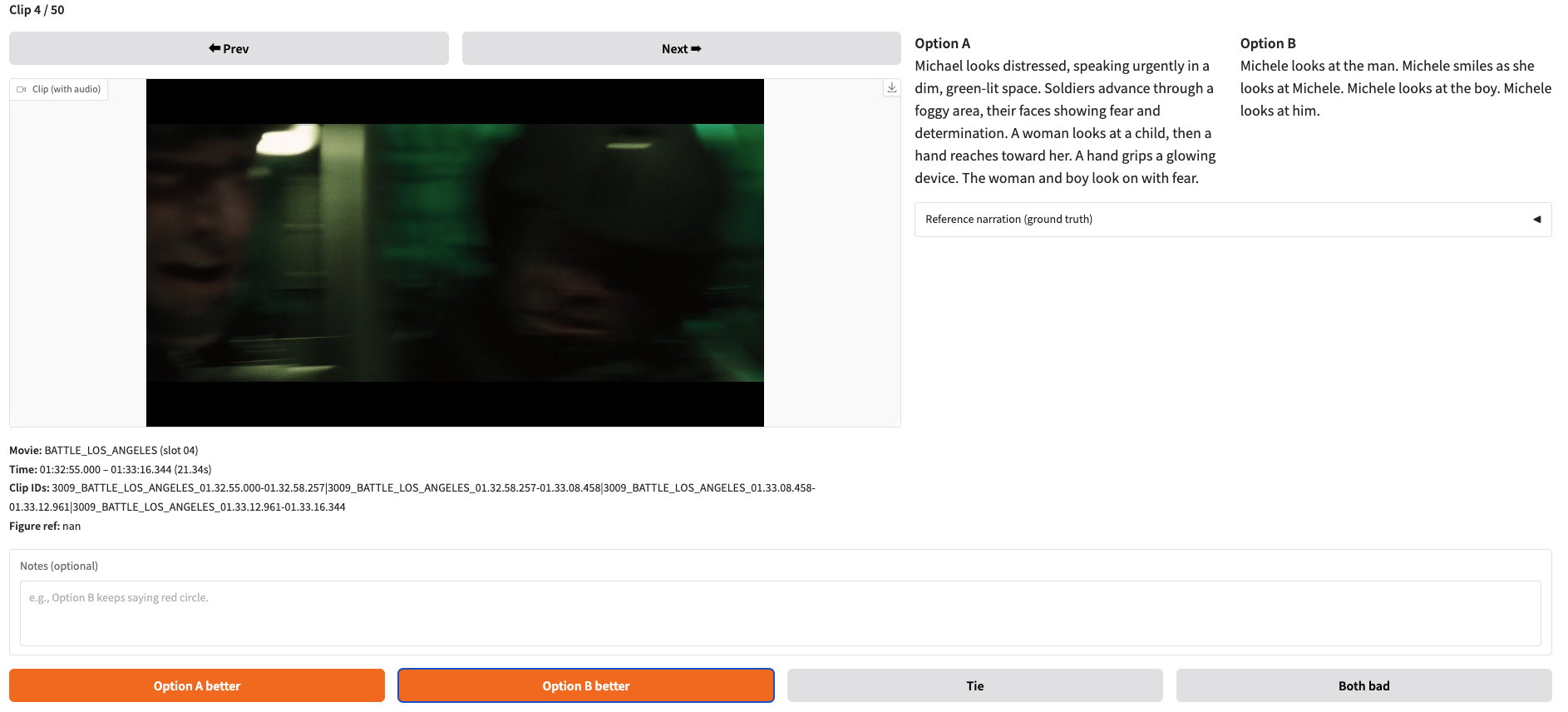}
\caption{
Human evaluation interface used in the study. Annotators watch a video clip and compare two candidate audio descriptions (Option A and Option B), whose order is randomized per trial. They select whether one description is better, whether the two are comparable (\emph{Tie}), or whether both descriptions are inadequate.
}
\label{fig:s8_interface}
\end{figure}

Figure~\ref{fig:s8_interface} illustrates the evaluation interface.
Annotators could replay the clip and optionally view the reference
audio narration (ground truth) for additional context, but no system
identifiers or auxiliary metadata were shown in order to maintain a
blind evaluation.

For each clip, annotators selected one of four options:

\begin{itemize}
\item \textbf{Option A better}: description A provides a better audio description.
\item \textbf{Option B better}: description B provides a better audio description.
\item \textbf{Tie}: both descriptions are of comparable quality.
\item \textbf{Both bad}: neither description adequately describes the scene.
\end{itemize}

Annotators were instructed to judge descriptions based on two criteria:
(i) how accurately the description reflects the visible content of the
clip, and (ii) whether the narration conveys the scene clearly and
coherently for a blind or low-vision audience. The study involved
three human evaluators, each rating all 50 clips.

\subsection{Results}
Table~\ref{tab:human_pref} summarizes the human preference results.
Across the 50 evaluated clips, annotators preferred StoryTeller
substantially more often than AutoAD-Zero, with StoryTeller receiving
76.0\% of the preferences compared to 5.8\% for the baseline while 9.1\% of the evaluations resulted ina tie and 9.1\% judged both descriptions inadequate.

In many cases, annotators selected StoryTeller when its descriptions
captured scene context or environmental cues visible in the frames,
whereas AutoAD-Zero descriptions tended to focus on shorter or more
generic actions.
\begin{table}[t]
\caption{Human preference comparison between StoryTeller and AutoAD-Zero.}
\label{tab:human_pref}
\centering
\begin{tabular}{l c}
\hline
Method & Preference (\%) \\
\hline
StoryTeller & 76.0 \\
AutoAD-Zero & 5.8 \\
Tie & 9.1 \\
Both bad & 9.1 \\
\hline
\end{tabular}
\end{table}

\section{Prompt Templates}

This section lists the prompt templates used in the StoryTeller pipeline, organized according to the modules described in Sec.~3 of the main paper: scene summarization, structured fact extraction, fact verification, and narration generation. The prompts guide the model to distinguish observed evidence from auxiliary cues such as public movie metadata. In particular, the fact-extraction schema records whether each candidate fact is suggested by visual, audio, or metadata input. Metadata may help propose candidates, such as possible character names, but these candidates are passed through a separate verification step before being added to memory or used in narration.

\subsection{Scene Summarization Prompt}
\textbf{Stage 1: Scene Summarization.}
This prompt generates a concise summary of the visual content of a single video clip. The goal of this stage is to produce a short, grounded description that captures the main observable event without
introducing speculative interpretation. The summary serves as the initial representation of the scene and is later expanded into structured narrative facts.

\input{Supplementary_Material/supp_src_codes/text_blocks/1_ST_prompt1}

\subsection{Structured Fact Extraction Prompt}
\textbf{Stage 2: Structured Fact Extraction.}
Given the scene summary and video clip, this prompt extracts structured facts describing observable actions, entities, and locations. The output follows a strict JSON schema in order to produce atomic, verifiable facts that can be independently checked. These structured facts form the basis of the narrative memory used
by StoryTeller.
\input{Supplementary_Material/supp_src_codes/text_blocks//2_ST_prompt_fact}

\subsection{Fact Verification Prompt}
\textbf{Stage 3: Fact Verification.}
To prevent hallucinated or weakly grounded observations from entering the memory, each extracted fact is passed through a verification step.
The model acts as a strict fact checker and determines whether the proposed fact is clearly supported by the visual evidence in the clip.
Only accepted facts are retained for downstream narration.
\input{Supplementary_Material/supp_src_codes/text_blocks//3_ST_prompt_verifier}

\subsection{Final Input Prompt}
\textbf{Stage 4: Final Narration Generation.}
The final prompt produces the audio description used for evaluation. It conditions on the verified fact set and optional narrative memory
context from previous clips. The prompt instructs the model to generate a concise, visually grounded narration that reflects the scene while
avoiding references to cameras or recording artifacts.
\input{Supplementary_Material/supp_src_codes/text_blocks//4_ST_final_narration}

\subsection{StoryAD-QA Question Generation Prompt}

\textbf{Track A (segment-only QA).}
The Track~A prompt asks the VLM to generate a multiple-choice question about the main visual event within a single clip. For question generation only, the VLM also receives a short Wikipedia movie overview as background. The prompt explicitly restricts the model to information that is directly visible, prohibits the use of dialogue, external knowledge, or speculative reasoning, and requires a short rationale grounded in visual evidence. If the clip does not contain a clear self-contained event, the model is instructed to output \texttt{SKIP}.

\input{Supplementary_Material/supp_src_codes/text_blocks//5_prompt_qa_track1}

\textbf{Track B (context-conditioned QA).}
The Track~B prompt introduces an explicit separation between a \emph{context window} and a \emph{target scene}. The context may introduce entities or situations, but the correct answer must be visually verifiable in the target scene. The prompt further discourages references to timestamps, dialogue, or outside knowledge to ensure questions remain grounded in observable visual information.

\input{Supplementary_Material/supp_src_codes/text_blocks//6_prompt_qa_track2}

\subsection{StoryAD-QA Question Answering Prompt}
During evaluation, we provide the language model with only textual inputs: the question and the corresponding audio descriptions for the scene. The model is instructed to select an answer using only the information contained in the AD, without the use of external knowledge, prior familiarity with the movie, or assumptions about events not described in the AD text. In addition, the model is required to produce a short rationale citing explicit evidence from the AD, encouraging answers that are grounded in the provided narration.

To reduce positional bias, the answer options are randomly shuffled before being presented to the language model during evaluation.

\input{Supplementary_Material/supp_src_codes/text_blocks//7_prompt_qa_answer}

%% file: Supplementary_Material/supp_src_codes/text_blocks/1_ST_prompt1.tex
\begin{tcolorbox}[breakable,colback=gray!5,colframe=gray!60,boxrule=0.4pt,arc=2pt]
\footnotesize

\textbf{Instruction}. You create concise, evidence-grounded clip summaries for audio description.
        Rules: present tense; one sentence; less than 25 words; 
        describe only what is clearly visible or audible; 
        no speculation, no spoilers; if identity is unclear, use neutral labels.
        
\textbf{Prompt}.  Summarize the clip in one concise sentence, describing only what is clearly visible
or audible.
\end{tcolorbox}

%% file: Supplementary_Material/supp_src_codes/text_blocks/2_ST_prompt_fact.tex
\begin{tcolorbox}[breakable,colback=gray!5,colframe=gray!60,boxrule=0.4pt,arc=2pt]
\footnotesize

\textbf{Instruction}. You extract structured, evidence-grounded facts from a video clip for audio description.
        Return ONLY valid JSON exactly matching the schema. No markdown. No extra keys. No inference.
        Keep facts atomic (one action each) and limit to 6 to 10 items. Use neutral labels if identity is unclear.

Fill the JSON schema below using only what is clearly visible or audible in the clip.

\textbf{Schema:}

\begin{verbatim}
{
  "scene_summary": "one sentence visual summary",
  "time_window": "start-end seconds or mm:ss-mm:ss",
  "facts": [
    {
      "subject": "who performed the action",
      "action": "verb phrase",
      "object": "what/who was acted upon; use empty string if 
      not needed",
      "location": "where in the scene",
      "time": "timestamp or short range",
      "evidence": "visual|audio|metadata",
      "confidence": "high|medium|low",
      "notes": "optional clarifications",
      "cluster_id": "if the action involves a TRACKLET_GUIDE entry,  
      copy its tracklet_id (e.g., trk_0003)",
      "character_id": "when 100% certain,     
      supply the canonical character_id from metadata"
    }
  ],
  "uncertain_observations": ["optional list describing ambiguities"]
}
\end{verbatim}

TRACKLET\_GUIDE (cluster IDs visible in this clip):  
\texttt{\{tracklet\_hint\}}

Spoiler-safe metadata snippets that may match this moment:  
\texttt{\{context\_snippets\}}

Metadata hints for this movie (use only if they match the video):  
\texttt{\{metadata\_hint\}}

\end{tcolorbox}

%% file: Supplementary_Material/supp_src_codes/text_blocks/3_ST_prompt_verifier.tex
\begin{tcolorbox}[breakable,colback=gray!5,colframe=gray!60,boxrule=0.4pt,arc=2pt]
\footnotesize

\textbf{Instruction}. You are a strict video-grounded fact checker. Accept only if the clip clearly supports the fact. 
        If ambiguous, inferred, off-screen, or not shown: reject.
        
\textbf{Prompt}.  You are verifying whether the FACT is truly visible or supported in this video clip. 
Use the video as the primary source. On the first line respond exactly with either ACCEPT or REJECT,
then provide a one-sentence justification citing what you observed.

FACT: \{fact\_text\}
\end{tcolorbox}

%% file: Supplementary_Material/supp_src_codes/text_blocks/4_ST_final_narration.tex
\begin{tcolorbox}[breakable,colback=gray!5,colframe=gray!60,boxrule=0.4pt,arc=2pt]
\footnotesize
\textbf{Instruction.}
You are an in-scene narrator describing events to blind or low-vision
listeners. Deliver the narration as a storyteller within the world of the
film. Never mention cameras, screens, or phrases such as ``in this video,''
``on screen,'' ``toward the camera,'' or ``looks at the camera.'' If a draft
would mention recording equipment, rewrite it so the line instead describes
how characters relate to one another or to their environment.

Rely on grounded details from the FACT\_BLOCK, keep the chronology clear,
and favor concise active sentences. When a FACT\_BLOCK entry includes a
\texttt{character\_id}, refer to that person by the provided first name only.
Do not invent or guess names. Use the provided first name when the character
is first mentioned; afterwards, pronouns such as he, she, or they may be used.

Highlight motion, intent, and key sounds. Omit filler and self-referential
phrasing. Keep the final description under 20 words. Use MEMORY\_CONTEXT
only when it directly supports continuity without speculation.

\medskip
\textbf{Clip duration:} \texttt{\{duration:.2f\}s}

\medskip
\textbf{Output format:}
Return JSON in the form
\texttt{\{"summarised\_AD": "...\,"\}}.

\medskip
\textbf{Examples.}
For a 0.8s clip:
\texttt{\{"summarised\_AD": "She looks at Riker."\}}

For a 1.4s clip:
\texttt{\{"summarised\_AD": "Paul looks at his wife lovingly."\}}

For a 2.6s clip:
\texttt{\{"summarised\_AD": "Returning to the room, Sarah peers into the darkness."\}}

For a 3.5s clip:
\texttt{\{"summarised\_AD": "Stephen looks at Sara as she walks to the door."\}}
\end{tcolorbox}

%% file: Supplementary_Material/supp_src_codes/text_blocks/5_prompt_qa_track1.tex
\begin{tcolorbox}[breakable,colback=gray!5,colframe=gray!60,boxrule=0.4pt,arc=2pt]
\footnotesize

You are given:\\
- A video clip from a movie.\\
- A movie overview for background context only.\\
Overview: \texttt{\{movie\_overview\}}\\

\textbf{IMPORTANT}: This clip may still be incomplete relative to the full movie.
Do NOT assume anything outside what is visible in this clip.
Do NOT use prior knowledge of the movie, characters, or story.
Use only what is directly shown. \\
\textbf{TASK}: Your task is to generate ONE 5-option multiple-choice question about the main visual event or progression shown in this clip. If the clip does NOT contain a clear, self-contained action, reaction, reveal, or cause–effect moment that is fully shown, output:\\
SKIP\\

The question MUST reflect the main development of the clip, not a small detail from a brief moment. Focus on the most important or climactic visible event.\\
Do NOT generate a question if:\\
- The moment is routine or transitional\\
- The answer depends on events outside this clip\\
- It requires guessing thoughts, intentions, or emotions\\
- It relies on what happens before or after the clip\\

If uncertain, output:\\
SKIP\\

Requirements:\\
- 1–2 sentence question\\
- Answerable using only this clip\\
- No “Why” questions\\
- No references to before or after the clip\\
- No psychological interpretation\\
- Focus on clearly visible actions or outcomes\\
All incorrect options must be plausible.\\

Output format:
\begin{verbatim}
Question:
<question>

A.
B.
C.
D.
E.

Correct Answer: <letter>

Rationale:
<1–2 sentences describing only visible evidence>
\end{verbatim}

\end{tcolorbox}

%% file: Supplementary_Material/supp_src_codes/text_blocks/6_prompt_qa_track2.tex
\begin{tcolorbox}[breakable,colback=gray!5,colframe=gray!60,boxrule=0.4pt,arc=2pt]
\footnotesize

You are given a movie video clip and a brief movie overview for background only.\\
Overview: \texttt{\{movie\_overview\}}\\

The first \texttt{\{context\_len\}} seconds of the clip are CONTEXT.\\
The remaining part of the clip is the TARGET SCENE.\\

The context may introduce people, objects, or locations that help interpret the target scene.

Generate ONE multiple-choice question (5 options A-E) that requires using the context to understand what is happening in the target scene.\\

\textbf{Key principle:} The answer must be visible in the TARGET SCENE, but the CONTEXT should help identify or interpret the object, person, or situation referenced in the question.

\textbf{Preferred patterns:}\\
- An object or person appears in the context, and the question asks about its state, action, or location in the target scene.\\
- Something in the target scene is ambiguous unless the viewer remembers the context.\\
- The context introduces an item or situation whose consequence or change becomes visible in the target scene.\\

Rules:\\
- The question must be answerable using only what is visibly shown in the video.
- The correct answer must be visually verifiable in the TARGET SCENE (after the first \texttt{\{context\_len\}} seconds).\\
- The context should help identify what the question refers to, but should not contain the answer itself.\\
- Do not rely on character names, dialogue, subtitles, or outside knowledge.\\
- Focus on observable actions, objects, interactions, or spatial relationships.\\
- Avoid trivial decorative details that do not affect the scene.\\
- Do not ask “Why” questions or infer emotions, intentions, or story meaning.\\
- Do not mention numeric timestamps, seconds, or "the first \texttt{\{context\_len\}} seconds".\\
- Do not refer to the video itself (no phrases like “in this clip” or “on screen”).\\

The question MUST reflect the main development of the clip, not a small detail from a brief moment. Focus on the most important or climactic visible event.\\

Do NOT generate a question if:\\
- The moment is routine or transitional\\
- The answer depends on events outside this clip\\
- It requires guessing thoughts, intentions, or emotions\\
- It relies on what happens before or after the clip\\

If there is no clear visual moment after the context that connects to something introduced earlier, output:\\
SKIP\\

Output format:
\begin{verbatim}
Question:
<question>

A. <option>
B. <option>
C. <option>
D. <option>
E. <option>

Correct Answer: <letter>

Rationale:
<1–2 sentences describing only the visible evidence in the target 
scene that confirms the answer>
\end{verbatim}

\end{tcolorbox}

%% file: Supplementary_Material/supp_src_codes/text_blocks/7_prompt_qa_answer.tex
\begin{tcolorbox}[breakable,colback=gray!5,colframe=gray!60,boxrule=0.4pt,arc=2pt]
\footnotesize
You are given:\\
- Audio Descriptions (AD) for a movie scene.\\
- A multiple-choice question about that scene.\\

Question:\\
\texttt{\{question\}}\\

Audio Descriptions (AD):\\
\texttt{\{ad\_text\}}\\

\textbf{IMPORTANT}: Use only information explicitly stated in the AD text.
Do NOT use prior knowledge of the movie.
Do NOT assume events outside the AD.
Do NOT guess characters' thoughts or intentions unless the AD explicitly states them.\\
\textbf{TASK}: Select the best answer (A, B, C, D, or E) based only on the AD.\\

Output format:
\begin{verbatim}
Answer: <letter>

Rationale:
<1–2 sentences citing only explicit AD evidence>
\end{verbatim}

\end{tcolorbox}

%% file: main.bib
@String(IJCV = {Int. J. Comput. Vis.})

@String(CVPR= {IEEE Conf. Comput. Vis. Pattern Recog.})

@String(ICCV= {Int. Conf. Comput. Vis.})

@String(ECCV= {Eur. Conf. Comput. Vis.})

@String(ACCV  = {ACCV})

@String(IJCV  = {IJCV})

@String(CVPR  = {CVPR})

@String(ICCV  = {ICCV})

@String(ECCV  = {ECCV})

@inproceedings{park2025narrad,
  title={NarrAD: Automatic Generation of Audio Descriptions for Movies with Rich Narrative Context},
  author={Park, Jaehyeong and Ye, Juncheol and Lee, Seungkook and Ka, Hyun W and Han, Dongsu},
  booktitle={Proceedings of the IEEE/CVF Winter Conference on Applications of Computer Vision (WACV)},
  pages={409--419},
  year={2025},
  organization={IEEE}
}

@inproceedings{zhang2024mm,
  title={Mm-narrator: Narrating Long-form Videos with Multimodal In-Context Learning},
  author={Zhang, Chaoyi and Lin, Kevin and Yang, Zhengyuan and Wang, Jianfeng and Li, Linjie and Lin, Chung-Ching and Liu, Zicheng and Wang, Lijuan},
  booktitle={Proceedings of the IEEE/CVF Conference on Computer Vision and Pattern Recognition (CVPR)},
  pages={13647--13657},
  year={2024}
}

@article{2023mmvid,
  author      = {Lin, Kevin and Ahmed, Faisal and Li, Linjie and Lin, Chung-Ching and Azarnasab, Ehsan and Yang, Zhengyuan and Wang, Jianfeng and Liang, Lin and Liu, Zicheng and Lu, Yumao and Liu, Ce and Wang, Lijuan},
  title       = {{MM-VID}: Advancing Video Understanding with GPT-4V(ision)},
  journal     = {arXiv preprint arXiv:2310.19773},
  year        = {2023},
  eprint      = {2310.19773},
  archivePrefix = {arXiv},
  primaryClass = {cs.CV},
}

@misc{Gemini3Flash,
  author = {Google},
  title = {{Gemini 3 Flash Preview} [Large language model]},
  year = {2026},
  url = {https://ai.google.dev/gemini-api/docs/models/gemini-3-flash-preview},
  note = {Accessed: June 29, 2026}
}

@inproceedings{autoadzero,
  title={{AutoAD-Zero}: A Training-Free Framework for Zero-Shot Audio Description},
  author={Xie, Junyu and Han, Tengda and Bain, Max and Nagrani, Arsha and Varol, G{\"u}l and Xie, Weidi and Zisserman, Andrew},
  booktitle={Proceedings of the Asian Conference on Computer Vision (ACCV)},
  pages={2265--2281},
  year={2024}
}

@InProceedings{autoad1,
    title={{AutoAD}: Movie Description in Context},  
    author={Tengda Han and Max Bain and Arsha Nagrani and G\"ul Varol and Weidi Xie and Andrew Zisserman},  
    booktitle={Proceedings of the IEEE/CVF Conference on Computer Vision and Pattern Recognition (CVPR)},  
    year={2023}}

@InProceedings{autoad2,
title={{AutoAD II: The Sequel} - Who, When, and What in Movie Audio Description},  
author={Tengda Han and Max Bain and Arsha Nagrani and G\"ul Varol and Weidi Xie and Andrew Zisserman},  
booktitle={Proceedings of the IEEE/CVF International Conference on Computer Vision (ICCV)},  
year={2023}}

@InProceedings{autoad3,
    author    = {Han, Tengda and Bain, Max and Nagrani, Arsha and Varol, G\"ul and Xie, Weidi and Zisserman, Andrew},
    title     = {AutoAD III: The Prequel - Back to the Pixels},
    booktitle = {Proceedings of the IEEE/CVF Conference on Computer Vision and Pattern Recognition (CVPR)},
    month     = {June},
    year      = {2024},
    pages     = {18164-18174}
}

@article{singh2025openai,
  title={Openai gpt-5 system card},
  author={Aaditya Singh and Adam Fry and Adam Perelman and Adam Tart and Adi Ganesh and Ahmed El-Kishky and Aidan McLaughlin and Aiden Low and AJ Ostrow and Akhila Ananthram and Akshay Nathan and Alan Luo and Alec Helyar and Aleksander Madry and Aleksandr Efremov and Aleksandra Spyra and Alex Baker-Whitcomb and Alex Beutel and Alex Karpenko and Alex Makelov and Alex Neitz and Alex Wei and Alexandra Barr and Alexandre Kirchmeyer and Alexey Ivanov and Alexi Christakis and Alistair Gillespie and Allison Tam and Ally Bennett and Alvin Wan and Alyssa Huang and Amy McDonald Sandjideh and Amy Yang and Ananya Kumar and Andre Saraiva and Andrea Vallone and Andrei Gheorghe and Andres Garcia Garcia and Andrew Braunstein and Andrew Liu and Andrew Schmidt and Andrey Mereskin and Andrey Mishchenko and Andy Applebaum and Andy Rogerson and Ann Rajan and Annie Wei and Anoop Kotha and Anubha Srivastava and Anushree Agrawal and Arun Vijayvergiya and Ashley Tyra and Ashvin Nair and Avi Nayak and Ben Eggers and Bessie Ji and Beth Hoover and Bill Chen and Blair Chen and Boaz Barak and Borys Minaiev and Botao Hao and Bowen Baker and Brad Lightcap and Brandon McKinzie and Brandon Wang and Brendan Quinn and Brian Fioca and Brian Hsu and Brian Yang and Brian Yu and Brian Zhang and Brittany Brenner and Callie Riggins Zetino and Cameron Raymond and Camillo Lugaresi and Carolina Paz and Cary Hudson and Cedric Whitney and Chak Li and Charles Chen and Charlotte Cole and Chelsea Voss and Chen Ding and Chen Shen and Chengdu Huang and Chris Colby and Chris Hallacy and Chris Koch and Chris Lu and Christina Kaplan and Christina Kim and CJ Minott-Henriques and Cliff Frey and Cody Yu and Coley Czarnecki and Colin Reid and Colin Wei and Cory Decareaux and Cristina Scheau and Cyril Zhang and Cyrus Forbes and Da Tang and Dakota Goldberg and Dan Roberts and Dana Palmie and Daniel Kappler and Daniel Levine and Daniel Wright and Dave Leo and David Lin and David Robinson and Declan Grabb and Derek Chen and Derek Lim and Derek Salama and Dibya Bhattacharjee and Dimitris Tsipras and Dinghua Li and Dingli Yu and DJ Strouse and Drew Williams and Dylan Hunn and Ed Bayes and Edwin Arbus and Ekin Akyurek and Elaine Ya Le and Elana Widmann and Eli Yani and Elizabeth Proehl and Enis Sert and Enoch Cheung and Eri Schwartz and Eric Han and Eric Jiang and Eric Mitchell and Eric Sigler and Eric Wallace and Erik Ritter and Erin Kavanaugh and Evan Mays and Evgenii Nikishin and Fangyuan Li and Felipe Petroski Such and Filipe de Avila Belbute Peres and Filippo Raso and Florent Bekerman and Foivos Tsimpourlas and Fotis Chantzis and Francis Song and Francis Zhang and Gaby Raila and Garrett McGrath and Gary Briggs and Gary Yang and Giambattista Parascandolo and Gildas Chabot and Grace Kim and Grace Zhao and Gregory Valiant and Guillaume Leclerc and Hadi Salman and Hanson Wang and Hao Sheng and Haoming Jiang and Haoyu Wang and Haozhun Jin and Harshit Sikchi and Heather Schmidt and Henry Aspegren and Honglin Chen and Huida Qiu and Hunter Lightman and Ian Covert and Ian Kivlichan and Ian Silber and Ian Sohl and Ibrahim Hammoud and Ignasi Clavera and Ikai Lan and Ilge Akkaya and Ilya Kostrikov and Irina Kofman and Isak Etinger and Ishaan Singal and Jackie Hehir and Jacob Huh and Jacqueline Pan and Jake Wilczynski and Jakub Pachocki and James Lee and James Quinn and Jamie Kiros and Janvi Kalra and Jasmyn Samaroo and Jason Wang and Jason Wolfe and Jay Chen and Jay Wang and Jean Harb and Jeffrey Han and Jeffrey Wang and Jennifer Zhao and Jeremy Chen and Jerene Yang and Jerry Tworek and Jesse Chand and Jessica Landon and Jessica Liang and Ji Lin and Jiancheng Liu and Jianfeng Wang and Jie Tang and Jihan Yin and Joanne Jang and Joel Morris and Joey Flynn and Johannes Ferstad and Johannes Heidecke and John Fishbein and John Hallman and Jonah Grant and Jonathan Chien and Jonathan Gordon and Jongsoo Park and Jordan Liss and Jos Kraaijeveld and Joseph Guay and Joseph Mo and Josh Lawson and Josh McGrath and Joshua Vendrow and Joy Jiao and Julian Lee and Julie Steele and Julie Wang and Junhua Mao and Kai Chen and Kai Hayashi and Kai Xiao and Kamyar Salahi and Kan Wu and Karan Sekhri and Karan Sharma and Karan Singhal and Karen Li and Kenny Nguyen and Keren Gu-Lemberg and Kevin King and Kevin Liu and Kevin Stone and Kevin Yu and Kristen Ying and Kristian Georgiev and Kristie Lim and Kushal Tirumala and Kyle Miller and Lama Ahmad and Larry Lv and Laura Clare and Laurance Fauconnet and Lauren Itow and Lauren Yang and Laurentia Romaniuk and Leah Anise and Lee Byron and Leher Pathak and Leon Maksin and Leyan Lo and Leyton Ho and Li Jing and Liang Wu and Liang Xiong and Lien Mamitsuka and Lin Yang and Lindsay McCallum and Lindsey Held and Liz Bourgeois and Logan Engstrom and Lorenz Kuhn and Louis Feuvrier and Lu Zhang and Lucas Switzer and Lukas Kondraciuk and Lukasz Kaiser and Manas Joglekar and Mandeep Singh and Mandip Shah and Manuka Stratta and Marcus Williams and Mark Chen and Mark Sun and Marselus Cayton and Martin Li and Marvin Zhang and Marwan Aljubeh and Matt Nichols and Matthew Haines and Max Schwarzer and Mayank Gupta and Meghan Shah and Melody Y. Guan and Melody Huang and Meng Dong and Mengqing Wang and Mia Glaese and Micah Carroll and Michael Lampe and Michael Malek and Michael Sharman and Michael Zhang and Michele Wang and Michelle Pokrass and Mihai Florian and Mikhail Pavlov and Miles Wang and Ming Chen and Mingxuan Wang and Minnia Feng and Mo Bavarian and Molly Lin and Moose Abdool and Mostafa Rohaninejad and Nacho Soto and Natalie Staudacher and Natan LaFontaine and Nathan Marwell and Nelson Liu and Nick Preston and Nick Turley and Nicklas Ansman and Nicole Blades and Nikil Pancha and Nikita Mikhaylin and Niko Felix and Nikunj Handa and Nishant Rai and Nitish Keskar and Noam Brown and Ofir Nachum and Oleg Boiko and Oleg Murk and Olivia Watkins and Oona Gleeson and Pamela Mishkin and Patryk Lesiewicz and Paul Baltescu and Pavel Belov and Peter Zhokhov and Philip Pronin and Phillip Guo and Phoebe Thacker and Qi Liu and Qiming Yuan and Qinghua Liu and Rachel Dias and Rachel Puckett and Rahul Arora and Ravi Teja Mullapudi and Raz Gaon and Reah Miyara and Rennie Song and Rishabh Aggarwal and RJ Marsan and Robel Yemiru and Robert Xiong and Rohan Kshirsagar and Rohan Nuttall and Roman Tsiupa and Ronen Eldan and Rose Wang and Roshan James and Roy Ziv and Rui Shu and Ruslan Nigmatullin and Saachi Jain and Saam Talaie and Sam Altman and Sam Arnesen and Sam Toizer and Sam Toyer and Samuel Miserendino and Sandhini Agarwal and Sarah Yoo and Savannah Heon and Scott Ethersmith and Sean Grove and Sean Taylor and Sebastien Bubeck and Sever Banesiu and Shaokyi Amdo and Shengjia Zhao and Sherwin Wu and Shibani Santurkar and Shiyu Zhao and Shraman Ray Chaudhuri and Shreyas Krishnaswamy and Shuaiqi and Xia and Shuyang Cheng and Shyamal Anadkat and Simón Posada Fishman and Simon Tobin and Siyuan Fu and Somay Jain and Song Mei and Sonya Egoian and Spencer Kim and Spug Golden and SQ Mah and Steph Lin and Stephen Imm and Steve Sharpe and Steve Yadlowsky and Sulman Choudhry and Sungwon Eum and Suvansh Sanjeev and Tabarak Khan and Tal Stramer and Tao Wang and Tao Xin and Tarun Gogineni and Taya Christianson and Ted Sanders and Tejal Patwardhan and Thomas Degry and Thomas Shadwell and Tianfu Fu and Tianshi Gao and Timur Garipov and Tina Sriskandarajah and Toki Sherbakov and Tomek Korbak and Tomer Kaftan and Tomo Hiratsuka and Tongzhou Wang and Tony Song and Tony Zhao and Troy Peterson and Val Kharitonov and Victoria Chernova and Vineet Kosaraju and Vishal Kuo and Vitchyr Pong and Vivek Verma and Vlad Petrov and Wanning Jiang and Weixing Zhang and Wenda Zhou and Wenlei Xie and Wenting Zhan and Wes McCabe and Will DePue and Will Ellsworth and Wulfie Bain and Wyatt Thompson and Xiangning Chen and Xiangyu Qi and Xin Xiang and Xinwei Shi and Yann Dubois and Yaodong Yu and Yara Khakbaz and Yifan Wu and Yilei Qian and Yin Tat Lee and Yinbo Chen and Yizhen Zhang and Yizhong Xiong and Yonglong Tian and Young Cha and Yu Bai and Yu Yang and Yuan Yuan and Yuanzhi Li and Yufeng Zhang and Yuguang Yang and Yujia Jin and Yun Jiang and Yunyun Wang and Yushi Wang and Yutian Liu and Zach Stubenvoll and Zehao Dou and Zheng Wu and Zhigang Wang},

  journal={arXiv preprint arXiv:2601.03267},
  year={2025}
}

@article{wang2025internvl3,
  title={Internvl3. 5: Advancing open-source multimodal models in versatility, reasoning, and efficiency},
  author={Weiyun Wang and Zhangwei Gao and Lixin Gu and Hengjun Pu and Long Cui and Xingguang Wei and Zhaoyang Liu and Linglin Jing and Shenglong Ye and Jie Shao and Zhaokai Wang and Zhe Chen and Hongjie Zhang and Ganlin Yang and Haomin Wang and Qi Wei and Jinhui Yin and Wenhao Li and Erfei Cui and Guanzhou Chen and Zichen Ding and Changyao Tian and Zhenyu Wu and Jingjing Xie and Zehao Li and Bowen Yang and Yuchen Duan and Xuehui Wang and Zhi Hou and Haoran Hao and Tianyi Zhang and Songze Li and Xiangyu Zhao and Haodong Duan and Nianchen Deng and Bin Fu and Yinan He and Yi Wang and Conghui He and Botian Shi and Junjun He and Yingtong Xiong and Han Lv and Lijun Wu and Wenqi Shao and Kaipeng Zhang and Huipeng Deng and Biqing Qi and Jiaye Ge and Qipeng Guo and Wenwei Zhang and Songyang Zhang and Maosong Cao and Junyao Lin and Kexian Tang and Jianfei Gao and Haian Huang and Yuzhe Gu and Chengqi Lyu and Huanze Tang and Rui Wang and Haijun Lv and Wanli Ouyang and Limin Wang and Min Dou and Xizhou Zhu and Tong Lu and Dahua Lin and Jifeng Dai and Weijie Su and Bowen Zhou and Kai Chen and Yu Qiao and Wenhai Wang and Gen Luo},
  journal={arXiv preprint arXiv:2508.18265},
  year={2025}
}

@inproceedings{huang2020movienet,
  author    = {Qingqiu Huang and Yu Xiong and Anyi Rao and Jiaze Wang and Dahua Lin},
  title     = {MovieNet: A Holistic Dataset for Movie Understanding},
  booktitle = {European Conference on Computer Vision (ECCV)},
  year      = {2020}
}

@inproceedings{vicol2018moviegraphs,
  author    = {Paul Vicol and Makarand Tapaswi and Lluis Castrejon and Sanja Fidler},
  title     = {MovieGraphs: Towards Understanding Human-Centric Situations from Videos},
  booktitle = {IEEE/CVF Conference on Computer Vision and Pattern Recognition (CVPR)},
  year      = {2018},
  doi       = {10.1109/CVPR.2018.00895}
}

@InProceedings{Soldan_2022_CVPR,
    author    = {Soldan, Mattia and Pardo, Alejandro and Alc\'azar, Juan Le\'on and Caba, Fabian and Zhao, Chen and Giancola, Silvio and Ghanem, Bernard},
    title     = {MAD: A Scalable Dataset for Language Grounding in Videos From Movie Audio Descriptions},
    booktitle = {Proceedings of the IEEE/CVF Conference on Computer Vision and Pattern Recognition (CVPR)},
    month     = {June},
    year      = {2022},
    pages     = {5026-5035}
}

@inproceedings{lei2018tvqa,
  title={Tvqa: Localized, compositional video question answering},
  author={Lei, Jie and Yu, Licheng and Bansal, Mohit and Berg, Tamara},
  booktitle={Proceedings of the Conference on Empirical Methods in Natural Language Processing (EMNLP)},
  pages={1369--1379},
  year={2018}
}

@inproceedings{papineni2002bleu,
    title = "{B}leu: a Method for Automatic Evaluation of Machine Translation",
    author = "Papineni, Kishore  and
      Roukos, Salim  and
      Ward, Todd  and
      Zhu, Wei-Jing",
    editor = "Isabelle, Pierre  and
      Charniak, Eugene  and
      Lin, Dekang",
    booktitle = "Proceedings of the 40th Annual Meeting of the Association for Computational Linguistics",
    month = jul,
    year = "2002",
    address = "Philadelphia, Pennsylvania, USA",
    publisher = "Association for Computational Linguistics",
    url = "https://aclanthology.org/P02-1040/",
    doi = "10.3115/1073083.1073135",
    pages = "311--318"
}

@inproceedings{linrouge,
    title = "{ROUGE}: A Package for Automatic Evaluation of Summaries",
    author = "Lin, Chin-Yew",
    booktitle = "Text Summarization Branches Out",
    month = jul,
    year = "2004",
    address = "Barcelona, Spain",
    publisher = "Association for Computational Linguistics",
    url = "https://aclanthology.org/W04-1013/",
    pages = "74--81"
}

@inproceedings{anderson2016spice,
  title={Spice: Semantic propositional image caption evaluation},
  author={Anderson, Peter and Fernando, Basura and Johnson, Mark and Gould, Stephen},
  booktitle={European Conference on Computer Vision (ECCV)},
  year={2016},
  organization={Springer}
}

@inproceedings{yue2023movie101,
  title={Movie101: A new movie understanding benchmark},
  author={Yue, Zihao and Zhang, Qi and Hu, Anwen and Zhang, Liang and Wang, Ziheng and Jin, Qin},
  booktitle={Proceedings of the Association for Computational Linguistics (ACL)},
  pages={4669--4684},
  year={2023}
}

@inproceedings{vedantam2015cider,
  title={{CIDEr}: Consensus-Based Image Description Evaluation},
  author={Vedantam, Ramakrishna and Zitnick, C. Lawrence and Parikh, Devi},
  booktitle={Proceedings of the IEEE/CVF Conference on Computer Vision and Pattern Recognition (CVPR)},
  pages={4566--4575},
  year={2015}
}

@inproceedings{tapaswi2016movieqa,
  author    = {Makarand Tapaswi and Yukun Zhu and Rainer Stiefelhagen and Antonio Torralba and Raquel Urtasun and Sanja Fidler},
  title     = {MovieQA: Understanding Stories in Movies Through Question-Answering},
  booktitle = {Proceedings of the IEEE/CVF Conference on Computer Vision and Pattern Recognition (CVPR)},
  year      = {2016},
  pages     = {4631--4640}
}

@article{rohrbach2017movie,
  author    = {Anna Rohrbach and Atousa Torabi and Marcus Rohrbach and Niket Tandon and Christopher Pal and Hugo Larochelle and Aaron Courville and Bernt Schiele},
  title     = {Movie Description},
  journal   = {International Journal of Computer Vision (IJCV)},
  year      = {2017},
  volume    = {123},
  number    = {1},
  pages     = {94--120},
  doi       = {10.1007/s11263-016-0987-1}
}

@article{gao2023rag_survey,
  author    = {Yunfan Gao and Yun Xiong and Xinyu Gao and Kangxiang Jia and Jinliu Pan and Yuxi Bi and Yi Dai and Jiawei Sun and Meng Wang and Haofen Wang},
  title     = {Retrieval-Augmented Generation for Large Language Models: A Survey},
  journal   = {arXiv preprint arXiv:2312.10997},
  year      = {2023},
  eprint    = {2312.10997},
  archivePrefix = {arXiv},
  primaryClass = {cs.CL}
}

@article{huang2024hallucination,
  author    = {Lei Huang and Weijiang Yu and Weitao Ma and Weihong Zhong and Zhangyin Feng and Haotian Wang and Qianglong Chen and Weihua Peng and Xiaocheng Feng and Bing Qin and Ting Liu},
  title     = {A Survey on Hallucination in Large Language Models: Principles, Taxonomy, Challenges, and Open Questions},
  journal   = {ACM Transactions on Information Systems},
  year      = {2024},
  volume    = {1},
  number    = {1},
  articleno = {1},
  month     = {January},
  doi       = {10.1145/3703155}
}

@article{long2025m3agent,
  author    = {Lin Long and Yichen He and Wentao Ye and Yiyuan Pan and Yuan Lin and Hang Li and Junbo Zhao and Wei Li},
  title     = {Seeing, Listening, Remembering, and Reasoning: A Multimodal Agent with Long-Term Memory},
  journal   = {arXiv preprint arXiv:2508.09736},
  year      = {2025},
  eprint    = {2508.09736},
  archivePrefix = {arXiv},
  primaryClass = {cs.CV},
  doi       = {10.48550/arXiv.2508.09736}
}

@article{he2024storyteller,
  author        = {Yichen He and Yuan Lin and Jianchao Wu and Hanchong Zhang and Yuchen Zhang and Ruicheng Le},
  title         = {StoryTeller: Improving Long Video Description through Global Audio-Visual Character Identification},
  journal       = {arXiv preprint arXiv:2411.07076},
  year          = {2024},
  eprint        = {2411.07076},
  archivePrefix = {arXiv},
  primaryClass  = {cs.CV},
  doi           = {10.48550/arXiv.2411.07076}
}

@article{wang2025videoem,
  author        = {Yun Wang and Long Zhang and Jingren Liu and Jiaqi Yan and Zhanjie Zhang and Jiahao Zheng and Ao Ma and Run Ling and Xun Yang and Dapeng Wu and Xiangyu Chen and Xuelong Li},
  title         = {Video-EM: Event-Centric Episodic Memory for Long-Form Video Understanding},
  journal       = {arXiv preprint arXiv:2508.09486},
  year          = {2025},
  eprint        = {2508.09486},
  archivePrefix = {arXiv},
  primaryClass  = {cs.CV},
  doi           = {10.48550/arXiv.2508.09486}
}

@article{luo2024videorag,
  title={Video-rag: Visually-aligned retrieval-augmented long video comprehension},
    author={Yongdong Luo and Xiawu Zheng and Guilin Li and Shukang Yin and Haojia Lin and Chaoyou Fu and Jinfa Huang and Jiayi Ji and Fei Chao and Jiebo Luo and Rongrong Ji},
  journal={Advances in Neural Information Processing Systems (NeurIPS)},
  volume={38},
  pages={168008--168033},
  year={2026}
}

@InProceedings{song2024moviechat,
    author    = {Song, Enxin and Chai, Wenhao and Wang, Guanhong and Zhang, Yucheng and Zhou, Haoyang and Wu, Feiyang and Chi, Haozhe and Guo, Xun and Ye, Tian and Zhang, Yanting and Lu, Yan and Hwang, Jenq-Neng and Wang, Gaoang},
    title     = {MovieChat: From Dense Token to Sparse Memory for Long Video Understanding},
    booktitle = {Proceedings of the IEEE/CVF Conference on Computer Vision and Pattern Recognition (CVPR)},
    month     = {June},
    year      = {2024},
    pages     = {18221-18232}
}

@inproceedings{fan2024videoagent,
  title={Videoagent: A memory-augmented multimodal agent for video understanding},
  author={Fan, Yue and Ma, Xiaojian and Wu, Rujie and Du, Yuntao and Li, Jiaqi and Gao, Zhi and Li, Qing},
  booktitle={European Conference on Computer Vision (ECCV)},
  pages={75--92},
  year={2024},
  organization={Springer}
}

@article{li2024optimus,
  title={Optimus-1: Hybrid multimodal memory empowered agents excel in long-horizon tasks},
  author={Li, Zaijing and Xie, Yuquan and Shao, Rui and Chen, Gongwei and Jiang, Dongmei and Nie, Liqiang},
  journal={Advances in Neural Information Processing Systems (NeurIPS)},
  volume={37},
  pages={49881--49913},
  year={2024}
}

@inproceedings{raajesh2024micap,
  title={MICap: A Unified Model for Identity-Aware Movie Descriptions},
  author={Raajesh, Haran and Desanur, Naveen Reddy and Khan, Zeeshan and Tapaswi, Makarand},
  booktitle = {Proceedings of the IEEE/CVF Conference on Computer Vision and Pattern Recognition (CVPR)},
  year={2024}
}

@article{shao2024visual,
  title={Visual cot: Advancing multi-modal language models with a comprehensive dataset and benchmark for chain-of-thought reasoning},
  author={Shao, Hao and Qian, Shengju and Xiao, Han and Song, Guanglu and Zong, Zhuofan and Wang, Letian and Liu, Yu and Li, Hongsheng},
  journal={Advances in Neural Information Processing Systems (NeurIPS)},
  volume={37},
  pages={8612--8642},
  year={2024}
}

@inproceedings{lewis2021retrievalaugmentedgenerationknowledgeintensivenlp,
  title={Retrieval-Augmented Generation for Knowledge-Intensive {NLP} Tasks},
  author={Lewis, Patrick and Perez, Ethan and Piktus, Aleksandra and Petroni, Fabio and Karpukhin, Vladimir and Goyal, Naman and K{\"u}ttler, Heinrich and Lewis, Mike and Yih, Wen-tau and Rockt{\"a}schel, Tim and Riedel, Sebastian and Kiela, Douwe},
  booktitle={Advances in Neural Information Processing Systems (NeurIPS)},
  volume={33},
  pages={9459--9474},
  year={2020}
}

@InProceedings{guu2020realm,
  title = 	 {Retrieval Augmented Language Model Pre-Training},
  author =       {Guu, Kelvin and Lee, Kenton and Tung, Zora and Pasupat, Panupong and Chang, Mingwei},
  booktitle = 	 {Proceedings of the International Conference on Machine Learning (ICML)},
  pages = 	 {3929--3938},
  year = 	 {2020},
  volume = 	 {119},
  publisher =    {PMLR},
  url = 	 {https://proceedings.mlr.press/v119/guu20a.html},
  
}

@inproceedings{karpukhin2020dpr,
    title = "Dense Passage Retrieval for Open-Domain Question Answering",
    author = "Karpukhin, Vladimir  and
      Oguz, Barlas  and
      Min, Sewon  and
      Lewis, Patrick  and
      Wu, Ledell  and
      Edunov, Sergey  and
      Chen, Danqi  and
      Yih, Wen-tau",

    booktitle = "Proceedings of the Conference on Empirical Methods in Natural Language Processing (EMNLP)",
    year = "2020",
    publisher = "Association for Computational Linguistics",
    url = "https://aclanthology.org/2020.emnlp-main.550/",
    doi = "10.18653/v1/2020.emnlp-main.550",
    pages = "6769--6781",
   
}

@inproceedings{izacard2021fid,
  title={Leveraging passage retrieval with generative models for open domain question answering},
  author={Izacard, Gautier and Grave, Edouard},
  booktitle={Proceedings of the Conference of the European Chapter of the Association for Computational Linguistics (EACL) },
  pages={874--880},
  year={2021}
}

@inproceedings{deng2019arcface,
  title={Arcface: Additive angular margin loss for deep face recognition},
  author={Deng, Jiankang and Guo, Jia and Xue, Niannan and Zafeiriou, Stefanos},
  booktitle={Proceedings of the IEEE/CVF Conference on Computer Vision and Pattern Recognition (CVPR)},
  pages={4690--4699},
  year={2019}
}

@inproceedings{kala2025adqa,
    title = "What You See is What You Ask: Evaluating Audio Descriptions",
    author = "Kala, Divy  and
      Khandelwal, Eshika  and
      Tapaswi, Makarand",
    editor = "Christodoulopoulos, Christos  and
      Chakraborty, Tanmoy  and
      Rose, Carolyn  and
      Peng, Violet",
    booktitle = "Proceedings of the Conference on Empirical Methods in Natural Language Processing (EMNLP)",
    month = nov,
    year = "2025",
    address = "Suzhou, China",
    publisher = "Association for Computational Linguistics",
    url = "https://aclanthology.org/2025.emnlp-main.1199/",
    doi = "10.18653/v1/2025.emnlp-main.1199",
    pages = "23496--23518",
    ISBN = "979-8-89176-332-6",
   
}

@article{li2026qwen3,
  title={Qwen3-VL-Embedding and Qwen3-VL-Reranker: A Unified Framework for State-of-the-Art Multimodal Retrieval and Ranking},
  author={Li, Mingxin and Zhang, Yanzhao and Long, Dingkun and Chen, Keqin and Song, Sibo and Bai, Shuai and Yang, Zhibo and Xie, Pengjun and Yang, An and Liu, Dayiheng and Zhou, Jingren and Lin, Junyang},
  journal={arXiv preprint arXiv:2601.04720},
  year={2026},
  eprint={2601.04720},
  archivePrefix={arXiv},
  primaryClass={cs.CV}
}

@misc{videollama2,
  title={VideoLLaMA 2: Advancing Spatial-Temporal Modeling and Audio Understanding in Video-LLMs},
  author={Cheng, Zesen and Leng, Sicong and Zhang, Hang and Xin, Yifei and Li, Xin and Chen, Guanzheng and Zhu, Yongxin and Zhang, Wenqi and Luo, Ziyang and Zhao, Deli and Bing, Lidong},
  year={2024},
  eprint={2406.07476},
  archivePrefix={arXiv},
  primaryClass={cs.CV}
}

@inproceedings{clip,
  title={Learning Transferable Visual Models From Natural Language Supervision},
  author={Radford, Alec and Kim, Jong Wook and Hallacy, Chris and Ramesh, Aditya and Goh, Gabriel and Agarwal, Sandhini and Sastry, Girish and Askell, Amanda and Mishkin, Pamela and Clark, Jack and Krueger, Gretchen and Sutskever, Ilya},
  year={2021},
  booktitle={Proceedings of the International Conference on Machine Learning (ICML)},
  pages={8748--8763}
}
